\documentclass{article}
\pdfoutput=1
\usepackage{microtype}
\usepackage{graphicx}
\usepackage{subfigure}
\usepackage{tabularray}
\usepackage{booktabs} 

\usepackage{hyperref}

\usepackage[accepted]{icml2024}

\usepackage{amsmath}
\usepackage{amssymb}
\usepackage{mathtools}
\usepackage{amsthm}

\usepackage[capitalize,noabbrev]{cleveref}

\theoremstyle{plain}

\theoremstyle{definition}

\theoremstyle{remark}

\usepackage[textsize=tiny]{todonotes}

\icmltitlerunning{Advancing DRL Agents in Commercial Fighting Games: Training, Integration, and Agent-Human Alignment}

\begin{document}

\twocolumn[
\icmltitle{Advancing DRL Agents in Commercial Fighting Games: Training, Integration, and Agent-Human Alignment}



\icmlsetsymbol{equal}{*}

\begin{icmlauthorlist}
\icmlauthor{Chen Zhang}{ustc,tencent}
\icmlauthor{Qiang He}{UCAS}
\icmlauthor{Yuan Zhou}{tencent}
\icmlauthor{Elvis S. Liu}{tencent}
\icmlauthor{Hong Wang}{tencent}
\icmlauthor{Jian Zhao}{ustc}
\icmlauthor{Yang Wang}{ustc,KeyLab,Suzhou}
\end{icmlauthorlist}

\icmlaffiliation{ustc}{University of Science and Technology of China (USTC), Hefei, China.}
\icmlaffiliation{KeyLab}{Key Laboratory of Precision and Intelligent Chemistry, USTC, China}
\icmlaffiliation{Suzhou}{Suzhou Institute for Advanced Research, USTC, China}
\icmlaffiliation{tencent}{Tencent Games}
\icmlaffiliation{UCAS}{Institute of Automation, Chinese Academy of Sciences, China}

\icmlcorrespondingauthor{Yang Wang}{angyan@ustc.edu.cn}
\icmlcorrespondingauthor{Elvis S. Liu}{\mbox{elvissyliu@tencent.com}}

\icmlkeywords{Machine Learning, ICML}

\vskip 0.3in
]



\printAffiliationsAndNotice{} 

\begin{abstract}
Deep Reinforcement Learning (DRL) agents have demonstrated impressive success in a wide range of game genres. However, previous research primarily focuses on optimizing DRL competence rather than addressing the challenge of prolonged player interaction. In this paper, we propose a practical DRL agent system for fighting games, named \textit{Shūkai}, which has been successfully deployed to Naruto Mobile, a popular fighting game with over 100 million registered users. \textit{Shūkai} quantifies the state to enhance generalizability, introducing Heterogeneous League Training (HELT) to achieve balanced competence, generalizability, and training efficiency. Furthermore, \textit{Shūkai} implements specific rewards to align the agent's behavior with human expectations.
\textit{Shūkai}'s ability to generalize is demonstrated by its consistent competence across all characters, even though it was trained on only 13\% of them. Additionally, HELT exhibits a remarkable 22\% improvement in sample efficiency. \textit{Shūkai} serves as a valuable training partner for players in Naruto Mobile, enabling them to enhance their abilities and skills.
\end{abstract} 
\section{Introduction}
\label{intro}

Deep Reinforcement Learning (DRL) has demonstrated its capability in sequential decision-making, ranging from robotics~\citep{schulman2015high,sac}, dialogue systems~\citep{instructgpt,llama2}, and games~\citep{mnih2015human,silver2017mastering,vinyals2019grandmaster}. Games~\citep{mnih2015human,moravvcik2017deepstack,silver2017mastering,brown2018superhuman,vinyals2019grandmaster}, in particular, provide a cost-effective and natural environment for training DRL agents due to the ease of data generation. However, previous work~\citep{berner2019dota,jaderberg2019human,ye2020mastering} primarily focuses on enhancing agent competence to professional levels and often overlooks the necessity of improving the human player experience in commercial applications~\citep{csikszentmihalyi2000beyond,bakker2011flow,cheng2016influence}.

 In gaming, DRL's potential extends beyond achieving high competence. The integration of DRL agents as adversaries in Player versus Environment (PvE) settings promises a richer gaming experience by offering dynamic and unpredictable interactions~\citep{shafer2012causes}. However, this integration is remarkably challenging. The computational demands of training DRL agents in modern games with extensive character pools are formidable. For instance, training scenarios can escalate in complexity with the increase in character matchups, as seen in cases requiring substantial resources~\citep{silver2017mastering,vinyals2019grandmaster,ye2020towards}. Furthermore, the sole pursuit of surpassing human performance in DRL agents may not necessarily enhance the player experience and restricts the practical deployment of DRL agents in games~\citep{cheng2016influence,bakker2011flow}. For example, players with lower competitive skill levels may find it frustrating to face overly strong agents. The inability to overcome such opponents can rapidly diminish a player's interest in the game. Therefore, it is imperative to ensure that DRL agents exhibit varying levels of competitiveness, aligning with the diverse skill levels of players and providing an enjoyable gaming experience.

To address these challenges, this paper proposes a DRL agent system called \textit{Shūkai}, which is specifically designed for fighting games. \textit{Shūkai} utilizes a unified DRL model capable of managing a diverse roster of characters, thereby significantly reducing the complexity inherent in large-scale character sets. \textit{Shūkai} significantly reduces the complexity associated with extensive character rosters and keeps a promising generalization ability. To counter potential overfitting due to training on a subset of characters~\citep{bejani2021systematic,cobbe2019quantifying}, our approach includes measures to enhance the generalizability of the unified model. Notably, \textit{Shūkai} features Heterogeneous League Training (HELT), a novel method that amalgamates agents of diverse structures, broadening the policy space and achieving a balance between competitive performance (competence) and policy generalization. Aligning agent behavior with human expectations is achieved through meticulously designed rewards, promoting a human-like gaming experience. A behavior evaluation system, informed by expert insights, is utilized to identify and align the behavioral nuances of DRL agents with human players.

\textit{Shūkai} has been extensively evaluated and deployed in Naruto Mobile, a renowned fighting game featuring over 400 characters and attracting more than 100 million registered players. The generalization of \textit{Shūkai} is evidenced by its consistent competence across all characters, despite being trained on only 13\% of them. HELT also presents 22\% improvement in terms of sample efficiency. Moreover, \textit{Shūkai} serves as a training partner for players in Naruto Mobile, beneficial for players to improve their abilities. The human-agent alignment of \textit{Shūkai} has positively impacted player engagement, as evidenced by significant improvements in player retention rates through the A/B test. Specifically, the game has experienced a significant next-day retention rate of over 50\%, a 7-day retention rate of over 20\%, and a 30-day retention rate of over 10\%. It has also experienced a maximum growth rate of 5\% and an average growth rate of 4\%.  These enhancements in player retention hold significant importance for a commercial game with a user base in the millions. These improvements demonstrate the real-world impact of DRL and give an example of the ground of DRL.

In summary, our contributions are threefold. \textbf{i)} We present a practical DRL agent system, \textit{Shūkai}, introducing HELT to improve agents' ability to generalization, and design rewards that align agents with human-player. \textbf{ii)} \textit{Shūkai} demonstrates robust generalizability and improved training efficiency in Naruto Mobile. The interaction with human players and the alignment of agent behavior with player expectations are comprehensively validated. \textbf{iii)} This work marks a further step in the deployment of DRL systems in a large-scale commercial fighting game. To the best of our knowledge, this is the first example of a DRL system being deployed in a large-scale commercial fighting game.
\section{Preliminaries}
In this section, we first introduce the commercial fighting game Naruto Mobile and then discuss the problem formulation.
\vspace{-0.2cm}
\subsection{Naruto Mobile}
\begin{figure}[htb]
\vskip 0.1in
    \centering
    \includegraphics[width=\linewidth]{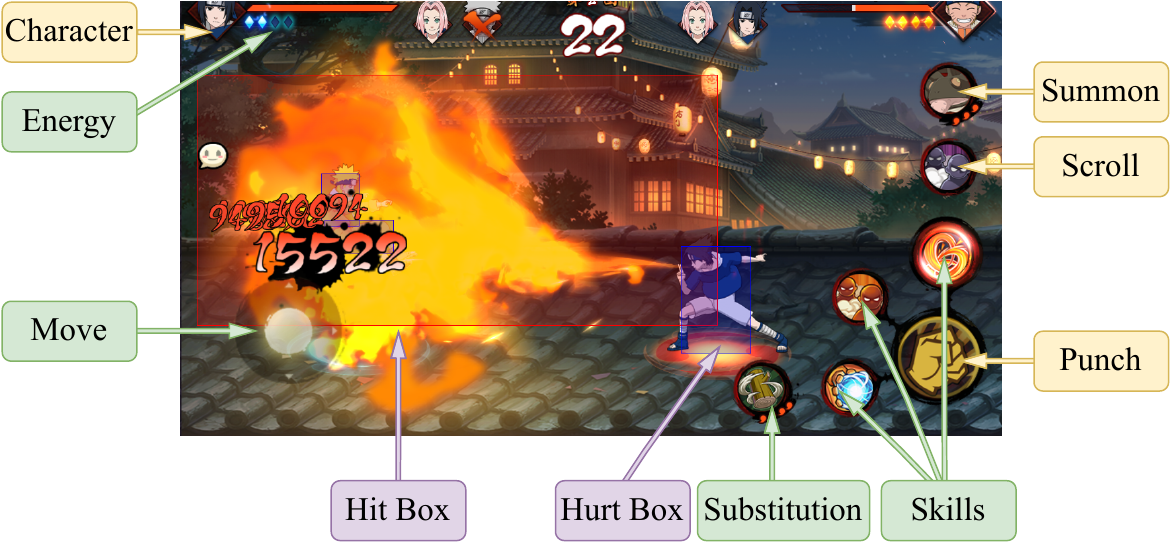}
    \caption{The interface of Naruto Mobile. The player selects a ninja from the character pool to fight against the opponent. The winning condition is to defeat the opponent. Each ninja has three skill buttons and a punch button, controlled by a virtual joystick. The hitbox and hurt box serve as fundamental game mechanics in Naruto Mobile, although they are not visible to the players. The substitution consumes energy and can be used to counter enemy attacks, creating opportunities for counterattacks. Scrolls and summons are additional skills with special effects, such as providing buffs.}
    \label{interface}
\vskip 0.1in
\end{figure}

Naruto Mobile is an online fighting game developed by Tencent Games with over 100 million registered users. Naruto Mobile has a large-scale character pool with more than 400 characters (ninjas). Each ninja has its special unique characteristics. \cref{interface} shows the interface of Naruto Mobile. Each episode of Naruto Mobile consists of two adversarial ninjas. 
Players of Naruto Mobile can choose a ninja from the character pool and use the ninja to fight against opponents. The winning condition for all players is to defeat their opponents, and the episode terminates when the condition is satisfied or the timeout. For more details of Naruto Mobile, readers can find in \cref{a_Naruto}.

Fighting games such as Naruto Mobile can be considered as a Markov decision process~\citep{sutton2018reinforcement}. At each time step, the agent records various pieces of information in their observation, including basic information, skill information, current skill information (referring to the activated skills), hit information (including the hitbox and hurt box), element information (describing entities with motion logic independent of characters, such as the behavior of skills), and buff information (referring to temporary enhancement effects). 

\begin{figure*}[htb]
    \vskip 0.1in
    \centering
    \includegraphics[width=\textwidth]{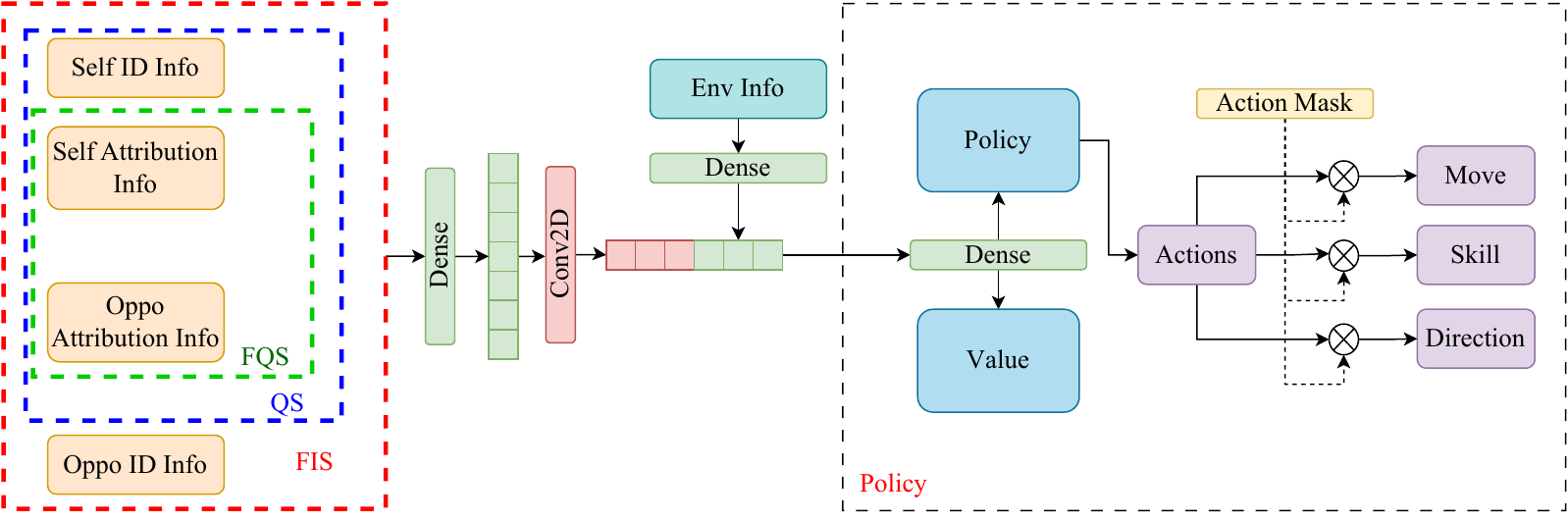}
    \vspace{-0.2in}
    \caption{The learning structure of FIS, QS, and FQS. FIS uses both ID and attribution information (numerical information) to model self and opponent. QS uses ID information to model self and only attribution information to model opponent. FQS models self and opponent only with attribution information. In this figure, the red dashed box contains FIS, the blue dashed box contains QS, and the green dashed box contains FQS. These features are processed by the network and then concatenated with the environment feature, the concatenated features are used to predict actions by the policy.}
    \label{fig_arche}
    \vskip 0.1in
\end{figure*}

In the state design, all this information is categorized into distinguishable categories to effectively model both the self and opponents (\cref{fig_arche}). This differentiation involves the utilization of ID information, which provides directed references to characters, skills, actions, etc., and attribution information, which represents the numerical values of entities (e.g., health points, cooldown values of skills, remaining time and location of the hurt box, etc.). The action space is categorized into movement, skills, and skill direction (specifically included to cater to skills that require specific orientations), ensuring the versatility of the model. For more details about the state, please refer to \cref{a_learning_arch}.
\subsection{Problem Formulation}
\label{markv}

Considering the opponent as a component of the environment, the MDP for Naruto Mobile's training agent is formalized as a six-element tuple $\langle \mathcal{S}, \mathcal{A}, P, R, T, \gamma \rangle$. Here, $\mathcal{S}$ represents the state space and $\mathcal{A}$ denotes the action space. $R$ is defined as the reward function, $P$ encapsulates the state transition probabilities, $T$ signifies the time horizon, and $\gamma$ is the discount factor. At each discrete time step $t$, the agent observes the current state of the environment, denoted as $s_t \in \mathcal{S}$. The agent's policy $\pi$, a mapping from states to actions, selects an action $a_t$ based on the current state, formalized as $a_t \sim \pi(\cdot|s_t)$. The state then transitions to the next state $s_{t+1} \sim P(\cdot|s_t,a_t)$, and the agent receives a reward $r_{t+1}$. 

The state value function $V$ and the action function $Q$ are always leveraged to optimize policy. They are defined as:
\begin{equation*}
\begin{split}
    &V_{\pi}(s)=\mathbb{E_{\pi}}[\sum\limits_{k=0}^{T}\gamma^{k}R_{t+k+1}|S_t=s],\\
    &Q_{\pi}(s)=\mathbb{E_{\pi}}[\sum\limits_{k=0}^{T}\gamma^{k}R_{t+k+1}|S_t=s, A_t=a].
\end{split}
\end{equation*}
The advantage function $A(s,a)=Q(s,a)-V(s)$ is used to measure the quality of action $a$ compared to the average quality.
\section{Method}
The application of the DRL agent system to Naruto Mobile presents several key challenges. Naruto Mobile features a large-scale character pool with over 400 characters, which is uncommon in other fighting games. While it may seem natural to assign a separate model to each character, deploying over 400 models in the game is unfeasible.

An intuitive approach is to use a unified model to control multiple characters, but this introduces another challenge. As the number of characters increases, the number of potential matchups between different characters also grows. In the case of 400 characters, the possible matchups reach up to $16\times10^4$, which is not favorable to training efficiency. To address this issue, one possible solution is to train on a subset of characters. However, this approach may lead to overfitting of the model to the subset~\citep{cobbe2019quantifying}, making the generalizability of the model decrease, resulting in the failure to adapt to unfamiliar characters who are out of the training subset. 

Therefore, the tricky challenge in designing the DRL agent system for Naruto Mobile lies in striking a balance between training efficiency, model competence, and generalization ability.

To address these challenges, \textit{Shūkai} applies different learning techniques such as heterogeneous neural network architecture, a unified model for controlling multiple characters, quantifying the network input to enhance the generalizability, and Heterogeneous League Training (HELT) to improve the training efficiency and competence.

\subsection{Heterogeneous Agents}
\label{heter agent}
To use a unified model to control multiple characters while generalizing to the characters out of the training subset, it is necessary to enable the network to learn more generalized knowledge rather than focusing on finding optimal responses within a subset of characters. A feasible approach is to imperfectize the network's input to enhance the model's generalization ability~\citep{schaul2015universal,dosovitskiy2016learning,finn2017model}.

As mentioned in \cref{markv}, in Narotu Mobile, all the information from the client is divided into two parts, the ID information directly refers to characters, skills, actions, etc., and the attribution information represents the numerical values of game entities. Based on ID information and attribution information, three heterogeneous networks for agents are employed:
\begin{itemize}
    \item \textbf{Full-ID-State (FIS):} Modeling self and opponent utilizing ID information and attribution information.
    \item \textbf{Quantitative-State (QS):} Modeling self with the ID information and attribution information while modeling the opponent only with the attribution information.
    \item \textbf{Full-Quantitative-state (FQS):} Modeling self and opponent utilizing only attribution information,  without any ID information. 
\end{itemize}
Quantitative states essentially refer to the utilization of incomplete information to construct representation. By employing quantitative states, the network shifts its focus from finding optimal decisions for specific actions of a designated character to learning knowledge about the game itself, thus enhancing its generalization ability. These agents with different structures will be applied in Heterogeneous League Training.
\subsection{Heterogeneous League Training}
To balance the competence and generalization ability of the DRL model, while considering training efficiency, \textit{Shūkai} introduce Heterogeneous League Training (HELT). HELT adopts a similar organizational structure to AlphaStar~\citep{vinyals2019grandmaster}, utilizing the main agent, main exploiter, and league exploiter to construct the league. The main agent serves as the primary training agent, tasked with defeating all opponents within the league. The main exploiter engages exclusively in battles against the contemporary main agent, with the primary objective of identifying weaknesses in the contemporary main agent. The league exploiter is employed to enhance the diversity within the league, ensuring that the main agent encounters a wide range of opponents.
\begin{figure}
\vskip -0.1in
    \centering
    \includegraphics[width=\linewidth]{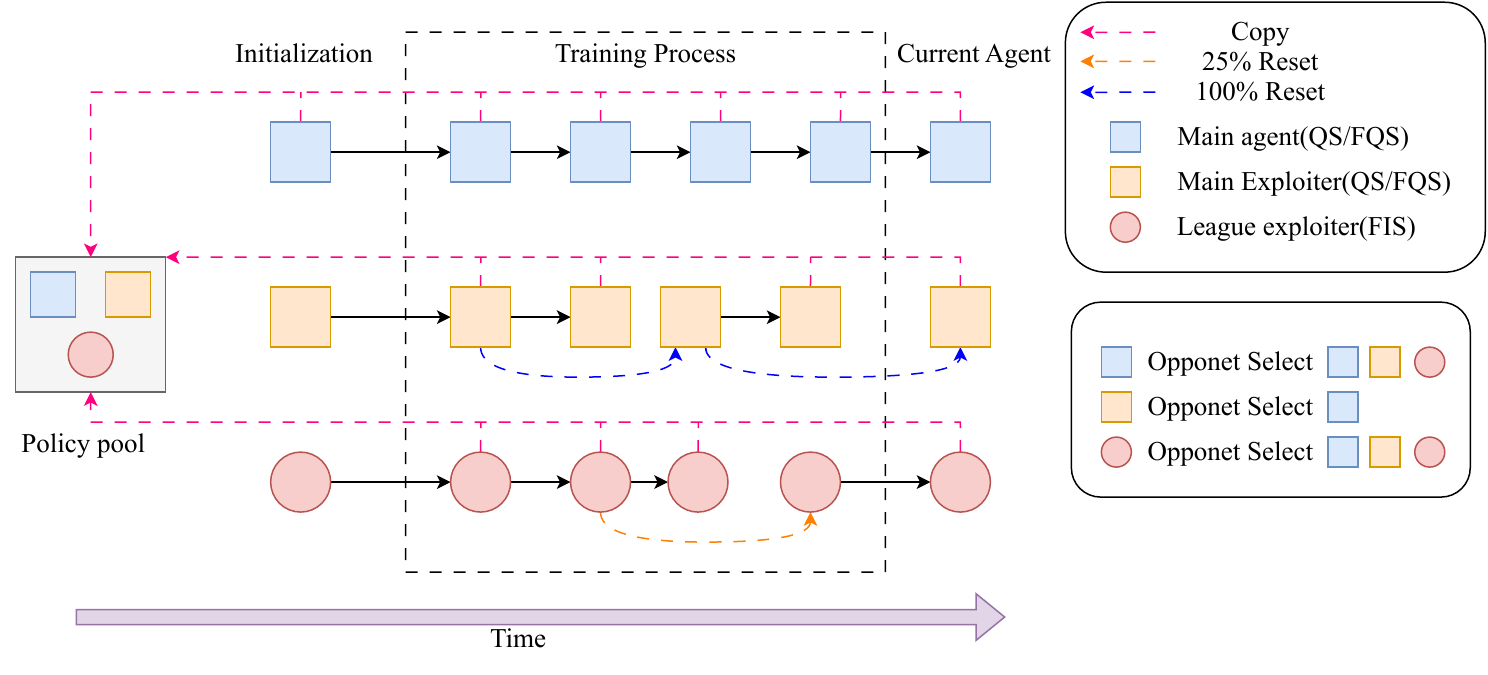}
    \vspace{-0.2in}
    \caption{Illustration of HELT. As time progresses, the main agent undergoes continuous training. Once it meets the win rate condition or reaches the timeout, a copy of the main agent is added to the policy pool. The main agent's objective is to defeat all opponents. Simultaneously, the main exploiter engages in battles with the main agent to discover its weaknesses. After meeting the win rate condition or reaching the timeout, the main exploiter is reset, and its copy is added to the policy pool. The league exploiter fights with all agents, and once the win rate condition or timeout is met, its copy is also added to the policy pool, with a 25\% probability of being reset. More details can find in \cref{a_learning_arch}}
    \label{fig_league_training_0}
\vskip -0.1in
\end{figure}

However, all the agents are homogeneous in AlphaStar, which may lead to overlapping exploration among agents within the league, resulting in reduced diversity. This reduction in diversity can have implications for the competence and generalization ability of the main agent and ultimately hinder the efficiency of training~\citep{lanctot2017unified}. In \textit{Shūkai}'s setting, the league of HELT consists of three distinct types of agents, differing primarily both in their \textbf{network structure} and in their \textbf{mechanism for selecting opponent mixture}. Different structures mentioned in \cref{heter agent} are used in HELT. 

The illustration of HELT is shown in \cref{fig_league_training_0}. Firstly, the QS agent or the FQS agent serves as the main agent to enhance generalizability. The main agents leverage the PFSP~\citep{vinyals2019grandmaster} mechanism that adapts the mixture probabilities proportionally to the win rate of each opponent against the agent. Secondly, the main exploiter shares the same structure (not the parameters) as the main agent and exclusively plays against the contemporary main agent. By engaging in battles against the contemporary main agent, the main exploiter assists in exploring overlooked aspects and helps the main agent uncover potential areas of improvement. Thirdly, the FIS agent serves as the league exploiter. The FIS agent utilizes ID information, enabling it to learn optimal responses for all characters within the training subset. As a result, it can quickly improve the competence of the league exploiter, thereby enhancing the overall training efficiency. When any agent achieves a winning rate of 80\% or encounters a timeout in the settings, it will load its copy into the league. After loading the current copy into the league, the main exploiter will reset, whereas the league exploiter has a 25\% probability of resetting. The main agent never reset.

Due to the structural differences between the league exploiter and the main agent, their exploration directions differ, thereby expanding the diversity within the policy space~\citep{lanctot2017unified}. This diversity better assists the main agent in learning generalized knowledge. Also, the FIS agent can quickly improve its competence cause it can find the best response to the characters in the training subset. Therefore, the league exploiter with the FIS structure will provide more valuable data to the main agent resulting in accelerating the training efficiency. The opponent policy selecting mechanism is PFSP~\citep{vinyals2019grandmaster}, readers can find the details in \cref{a_oppo_select}.

\subsection{Agent-Human Alignment}
\label{methods_align}
Deploying \textit{Shūkai} into Naruto Mobile presents another key challenge of aligning with human players' expectations. The goal of the DRL agent system is to assist players in improving their abilities, so it is essential to provide an agent that offers a certain level of challenge while still being beatable. To achieve this alignment, we introduce additional action masks to constrain the behavior of the DRL agent. In \textit{Shūkai}, we reduce the frequency of agent predictions to align with the reaction speed of human players. We also designed a mechanism that controls the execution probability of the action to generate different levels of \textit{Shūkai}.

To enhance the human-like behavior of the DRL agent, we collected a significant amount of player gameplay data and discovered that players of Naruto Mobile primarily fall into three archetypes $\langle balanced,cautious,aggressive\rangle $. Aggressive players tend to favor proactive attacks, cautious players lean towards defensive counterattacks, and balanced players adapt their actions based on the current situation. Based on these observations, we designed three different types of reward functions to align with the behavioral styles of human players.
\begin{itemize}
    \item \textbf{Balanced.} The agent will trade off the opponent's remaining health point, its health point, and the available resources. It aims to strategically defeat the opponent at the right moment and in a suitable manner.
    \item \textbf{Cautious.} The agent will prioritize its resources and remaining health points, aiming to minimize damage taken while winning the game.
    \item \textbf{Aggressive.} The agent will focus on the opponent's remaining health points. The objective of the agent is to defeat the opponent as quickly as possible.
\end{itemize}

These reward functions can be used as modular components and integrated into HELT.
By utilizing HELT with these reward functions, we can facilitate the development of multi-style agents. Also, we can prevent DRL agents from making confusing or seemingly illogical actions that players may find off-putting. This helps to maintain the illusion of human-like behavior and avoid detection by the players. Please check \cref{application} for the agent-human alignment.

\subsection{Policy Improvement}
In~\cref{markv}, by modeling the fighting game as a Markov process, we can leverage reinforcement learning algorithms to optimize the agent's strategies. We use the Proximal Policy Optimization (PPO) algorithm~\citep{schulman2017proximal} in our system. PPO trains a value function $V_{\phi}(s_t)$ with a policy $\pi_{\theta}(a_t|s_t)$, the utilization of importance sampling makes it highly feasible to apply reinforcement learning algorithms in distributed scenarios.

In the training process, the trajectories are sampled from multiple policies, which can exhibit significant deviations from the current policy $\pi_{\theta}$. Our updated objectives have remained consistent with PPO~\citep{schulman2017proximal}:
\begin{equation*}
     \mathcal{L}^{\pi}(\theta)=\mathbb{E}_{t}[min(\rho_{t}(\theta)\hat{A_t}, clip(\rho_t(\theta),1-\epsilon,1+\epsilon)\hat{A_t})], 
\end{equation*}
where $\rho_t(\theta)={\pi_{\theta}(s_t|a_t)}/{\pi_{old}(s_t|a_t)}$ is the importance sampling weight and $\hat{A_t}$ is the advantage estimation~\citep{sutton2018reinforcement} that is calculated by Generalized Advantage Estimation (GAE) \citep{schulman2015high}. More details can be found in \cref{a_policy}.

\begin{figure*}[htb]
\vskip 0.1in
	\centering
	\subfigure[Fight with familiar opponents] {\includegraphics[width=.33\textwidth]{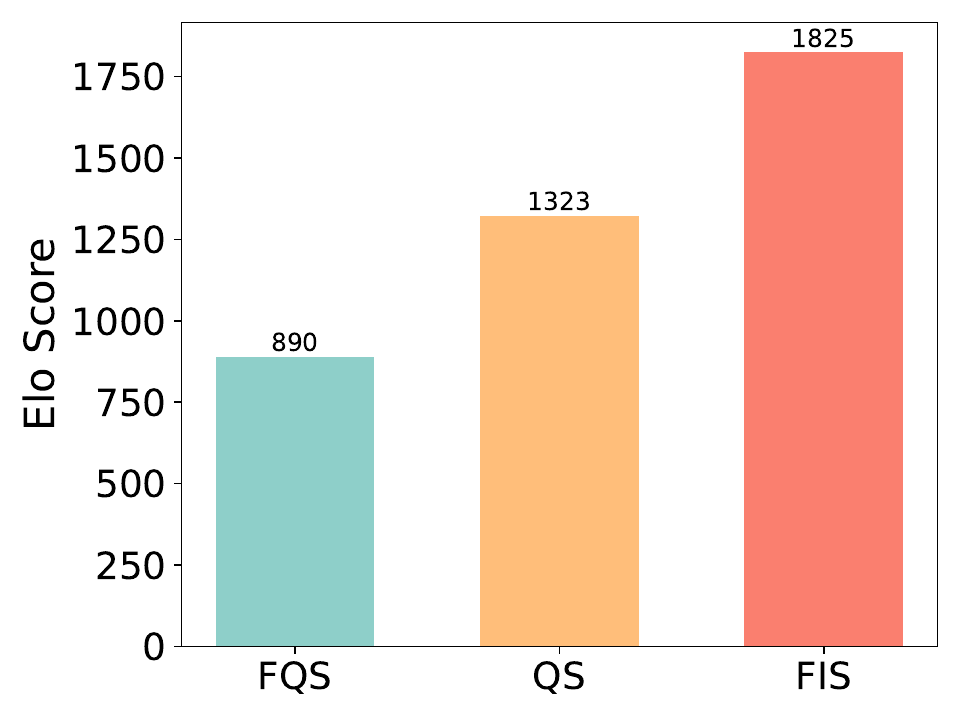}\label{performance}}
	\subfigure[Fight with unfamiliar opponents] {\includegraphics[width=.33\textwidth]{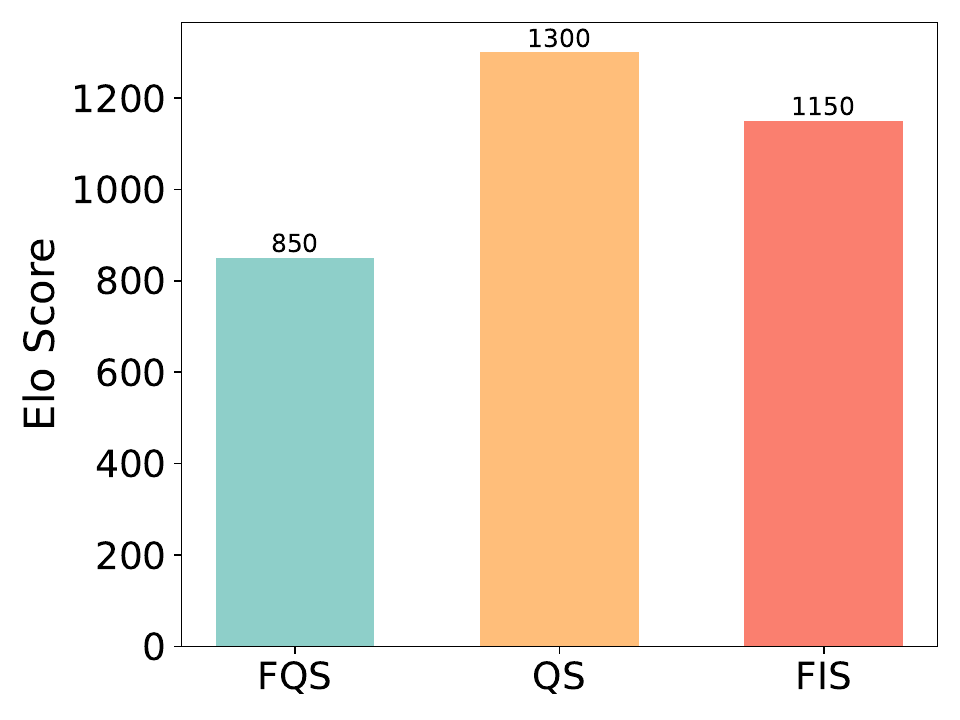}\label{generalization}}
	\subfigure[Training efficiency of HELT] {\includegraphics[width=.33\textwidth]{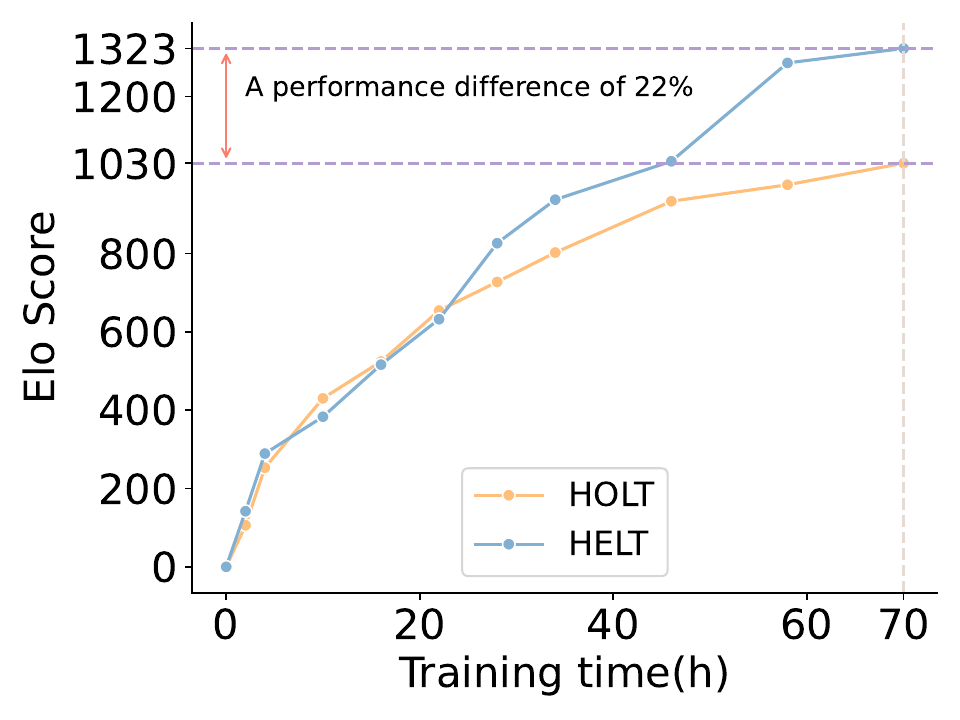}\label{Training_eff}}
	\caption{(a) Competence of different structures serve as the main agent in HELT, FIS agent outperforms other agents. (b) Generalizability of different structures serves as the main agent in HELT, the competence of the FIS agent dropped by 36\%, while the competence of the QS and FQS agents remained stable. (c) Training efficiency of HELT. After 70 hours of training, HELT achieved a 22\% improvement in competence compared to the QS network with homogeneous league training (HOLT).}
	\label{fig_ablation}
\vskip 0.1in
\end{figure*}
\section{Experiments}
\label{experiment}
\subsection{Experimental Setup}
In this section, we present the experimental results of HELT, using Naruto Mobile as the evaluation platform. For clarity, we define the HELT agent as the final version of the main agent, while the league agents consist of the main agent, main exploiter, and league exploiter. The terms `FIS agent', `QS agent', and `FQS agent' refer to the FIS, QS, and FQS network structures, respectively, when they serve as the main agent. The familiar characters refer to the characters who are in the training subset, and the unfamiliar characters refer to the characters who are out of the training subset.

Naruto Mobile has over 400 characters, making it impractical to train them all simultaneously. To manage this, we categorized the characters into four distinct levels $(S, A, B, C)$ based on their in-game strength and expert prior. We constructed a training subset by selecting 15 characters from the S-level, 15 from the A-level, 10 from the B-level, and 10 from the C-level. In total, 50 characters have been chosen across the four levels (13\% of all characters).

In our experimental setup, all agents were trained using 4 NVIDIA T4 GPUs and 3000 CPU cores. The league training consisted of a main agent, a main exploiter, and a league exploiter. A total of 12 GPUs and 9000 CPU cores were utilized for each league training session. The training process comprised 10 rounds, the first 7 rounds lasting 6 hours each, and the final 3 rounds lasting 12 hours each.

\subsection{Results}
To demonstrate the specific competence of HELT, we constructed an "oracle agent" that had access to all information, including cheat information from the client. Additionally, the oracle agent features an expanded character pool, allowing it to choose any character regardless of whether it was in HELT's character pool and training a longer time than HELT agents. HELT agents with different structures are evaluated by having them compete against the oracle agent, the competence results are presented in \cref{performance}, and the generalizability results are presented in \cref{generalization}. Each evaluation consisted of 2500 matches. The characters in the character pool were uniformly sampled (all the matchups have been sampled). The competence of the model can be represented by its Elo score~\citep{elo1978rating}, with a higher Elo score indicating a higher level of competence.

\textbf{Competence.} The competence result is shown in \cref{performance}. The FIS agents outperformed agents with other structures. This finding suggests that utilizing ID information to model the opponent can significantly enhance competence when dealing with familiar characters. By incorporating this approach, the agents can effectively learn and adapt to the unique characteristics and optimal strategies of different characters in the training subset, resulting in improved overall competence. 

In contrast, the FQS and QS agents, which quantify the states, did not perform satisfactorily when facing familiar characters. Particularly, the FQS network exhibited the worst competence. This suggests that quantifying the states can lead to a decrease in competence, and the higher the degree of quantization, the more pronounced the competence decline.

\textbf{Generalizability.} \cref{generalization} displays the result of generalizability, the results indicate an interesting phenomenon: when facing the unfamiliar characters, the competence of the FIS agent dropped by 36\%, while the competence of the QS and FQS agents remained stable, with negligible competence decline. This result can be attributed to the FIS agent's over-reliance on ID information for modeling different characters, which reduced its generalization capability. Conversely, appropriately quantifying the states allowed the QS and FQS agents to effectively learn the patterns of gameplay. The excellent generalization performance of QS and FQS agents is crucial in real-game scenarios where targeted training for all characters is not feasible. Therefore, when dealing with a large-scale character pool, a purely ID-based FIS structure may not be the optimal choice.

\textbf{Training Efficiency.} To compare the improvement in training efficiency brought by HELT, we trained a HELT agent with a QS network as the main agent and compared it with a QS network that constructs homogeneous league training~\citep{vinyals2019grandmaster}.
During the training process, the training resources and settings for both models are the same. Specifically, each agent had access to the same number of GPUs and CPUs and set the same number of training epochs and time limits to ensure a fair comparison. The result is shown in \cref{Training_eff}.
By comparing the training results, we observed a significant competence difference. After 70 hours of training, the HELT system achieved a 22\% improvement in competence compared to the QS network with homogeneous league training. This indicates that the HELT system can converge faster and reach a higher level of competence within the same training time.

The above experimental results indicate that by quantifying states, DRL agents can enhance generalization while maintaining competence. The introduction of HELT can improve training efficiency and enhance the competence of DRL agents, thereby mitigating key challenges in implementing DRL systems in real-world applications to games. Due to the considerations of both competence and generalization, the QS network has been chosen as the main agent for training in the practical deployment.

\section{Real-world application of Shūkai}
\label{application}
\subsection{Human Evaluation}
The main application of \textit{Shūkai} in the Naruto Mobile is to serve as a player's training partner. The primary purpose is to enhance the gaming abilities and skills of players through fighting with \textit{Shūkai}. Therefore, ensuring a satisfying player gaming experience is of paramount importance, and the competence of DRL agents needs to be challenging while still beatable~\citep{bakker2011flow}. We classify \textit{Shūkai} into three levels: beginner, intermediate, and advanced, based on the number of frames for delay prediction and the probability of action execution. The higher the level of the agent, the lower the delay in frame prediction and the higher the probability of executing actions.

\begin{figure}[htb]
    \centering
    \includegraphics[width=\linewidth]{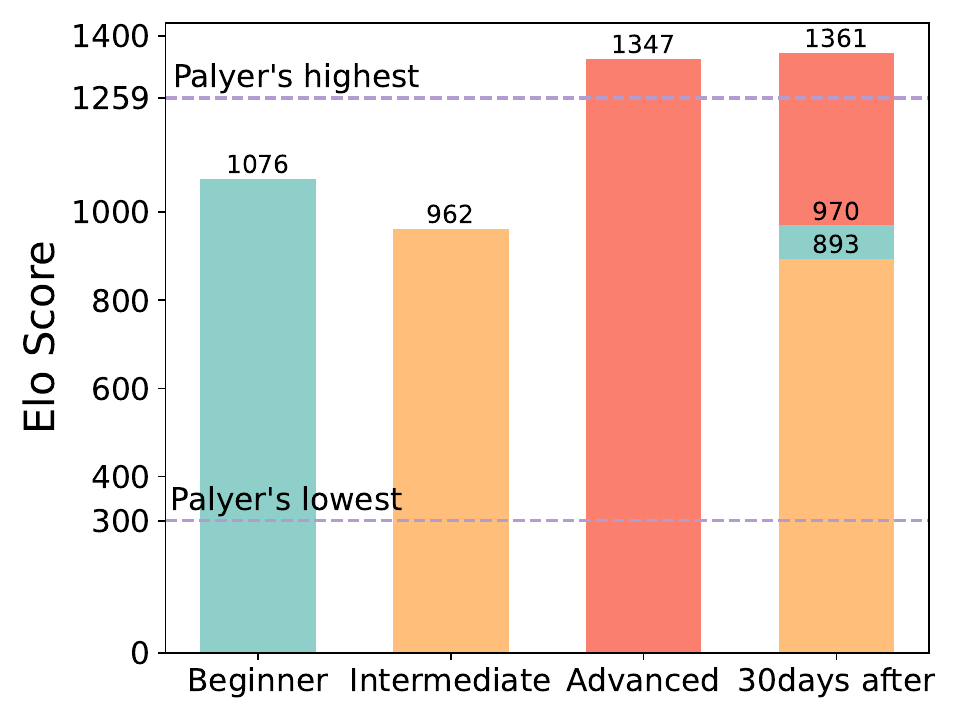}
    \vskip -0.1in
    \caption{The competence of three different levels of \textit{Shūkai} compete against human players in-game matches. The Elo scores of different level agents after 30 days are represented in the rightmost column. After 30 days, \textit{Shūkai} beginner and \textit{Shūkai} intermediate experience a decrease in Elo score, indicating \textit{Shūkai} can help players to enhance their abilities. \textit{Shūkai} advanced maintain its Elo score, suggesting advanced \textit{Shūkai} is challenging even for the skillful player.}
    \label{fig_against_human}
\vskip 0.1in
\end{figure}

The ranking system of Naruto Mobile uses Elo scores, ranging from 0 to 2599. \textit{Shūkai} are primarily deployed in the score range of 0-1299. Within this range, we deploy \textit{Shūkai} beginner for scores 0-600, \textit{Shūkai} intermediate for scores 600-999, and \textit{Shūkai} advanced for scores 999-1299. \textit{Shūkai} has been online in Naruto for over a year and has participated in more than 500 million player-agent matches. We conducted a competence analysis using a sample of 40 million game matches from player gameplay data, collected over time. The results are displayed in the \cref{fig_against_human}.

In each match, players have a certain probability of being matched with either \textit{Shūkai} or other human players as opponents. With players having different skill levels, talents, and experiences, using human players' performance against each other as a baseline could introduce certain inconsistencies. We ultimately chose to focus on \textit{Shūkai}'s Elo score change trend. This aim to showcase \textit{Shūkai}'s impact within specific score ranges during 30 days, rather than individual player performance changes.

According to the \cref{fig_against_human}, \textit{Shūkai} beginner achieved Elo scores surpassing their respective rating brackets. This can be attributed to the novice players' initial lack of proficiency which needs competence improvement. After 30 days, \textit{Shūkai} beginner experienced a 9\% decrease in Elo score, providing evidence of player skill enhancement through the utilization of \textit{Shūkai}. Similarly, \textit{Shūkai} intermediate followed a similar pattern to \textit{Shūkai} beginner. With the training provided by \textit{Shūkai} intermediate, players demonstrated an improvement in skill level for 30 days. Conversely, the Elo scores of the \textit{Shūkai} advanced increased after 30 days. However, the difference in Elo scores from the highest player rating was less than 10\%, suggesting that though \textit{Shūkai} advanced provided a challenge within that rating bracket, they were not unbeatable. The results also demonstrate the generalizability of \textit{Shūkai} due to the inclusion of characters selected by players that were not exclusively from the training subset which is the same as the experiment.

\begin{figure*}[htb]
\vskip 0.1in
\centering 
\subfigure[Substitution]  
{\includegraphics[width=.18\textwidth]{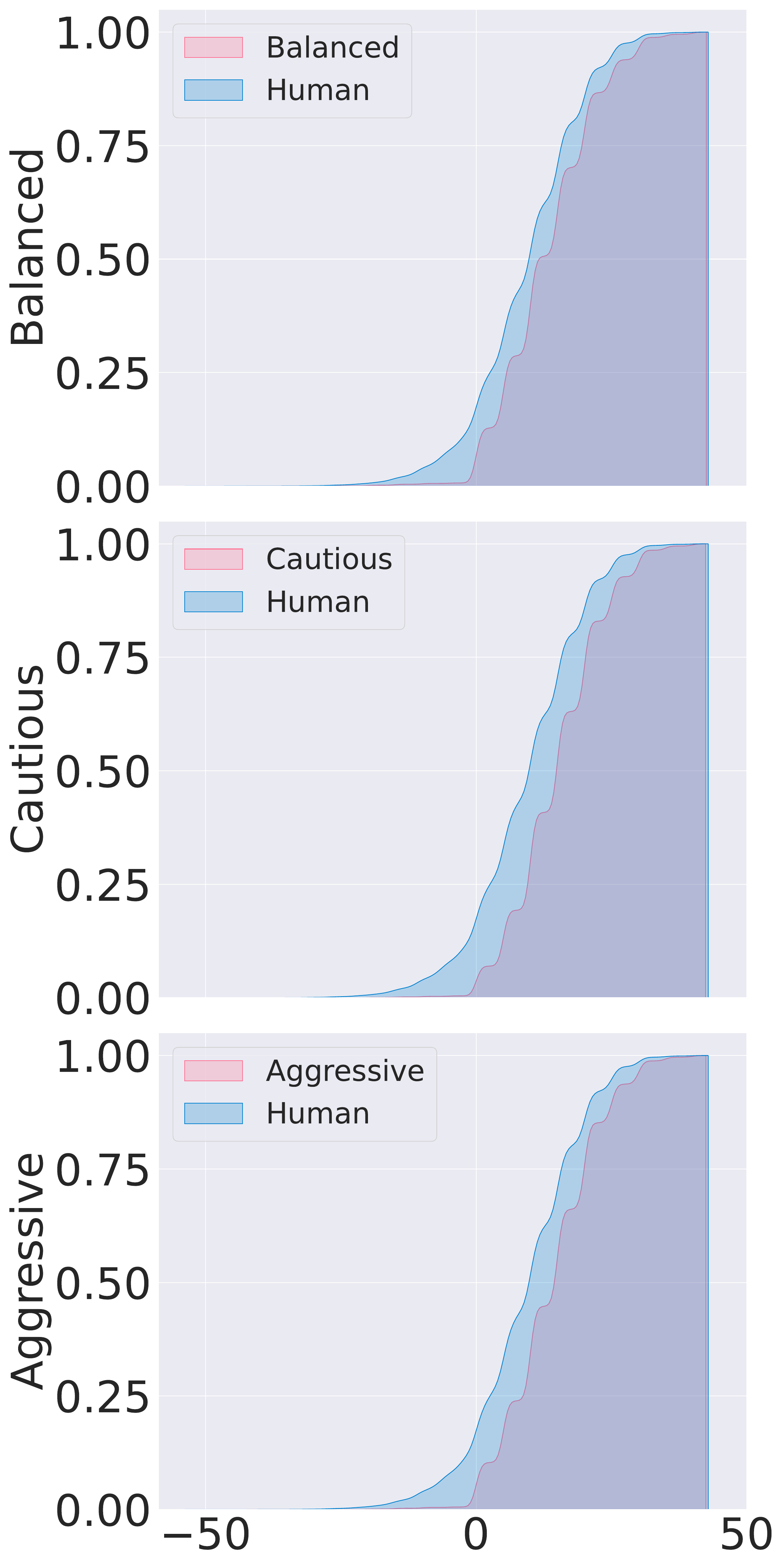}\label{Substitution}} \subfigure[Special] {\includegraphics[width=.18\textwidth]{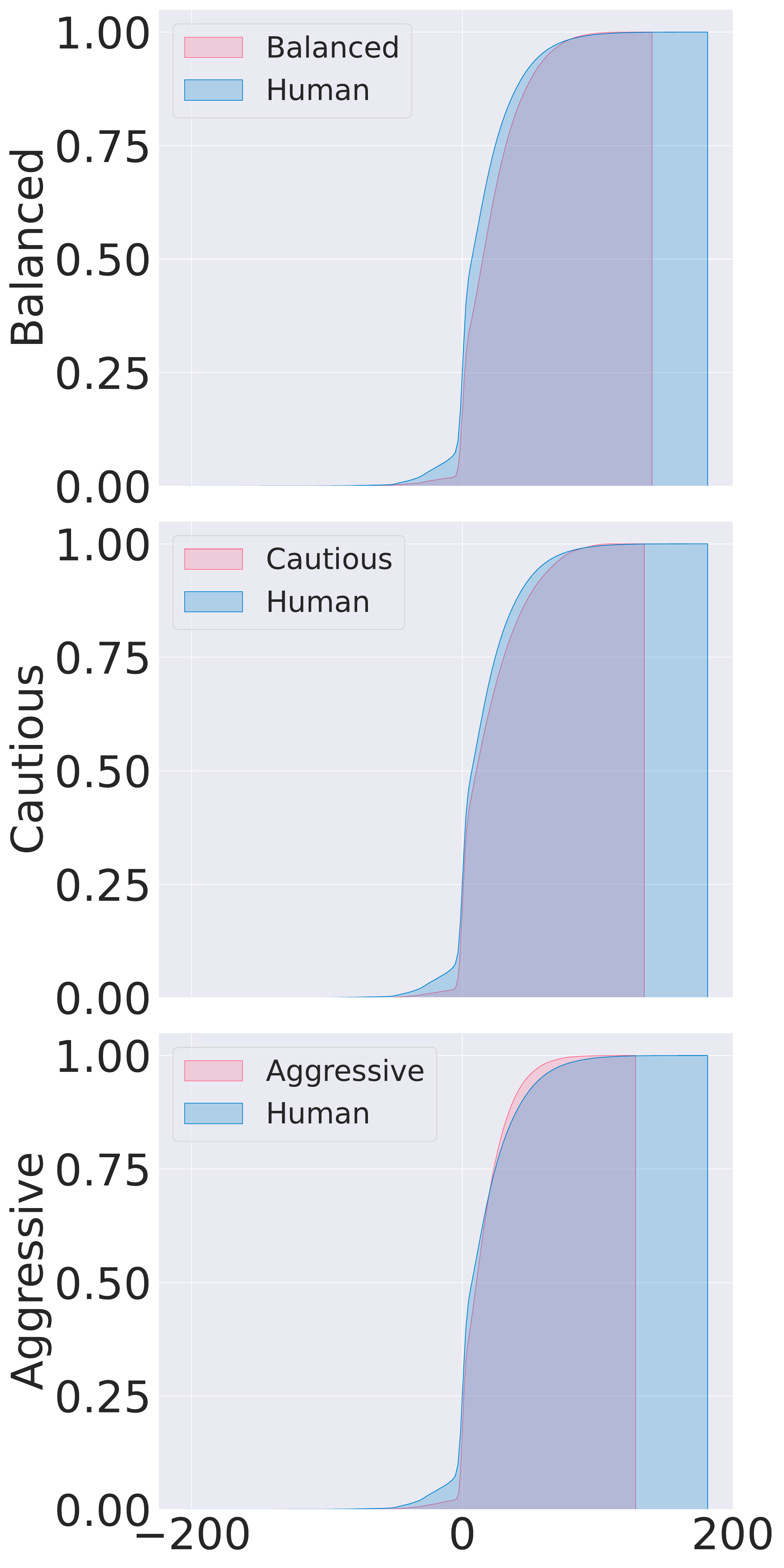}\label{Special}} \subfigure[Blitz] {\includegraphics[width=.18\textwidth]{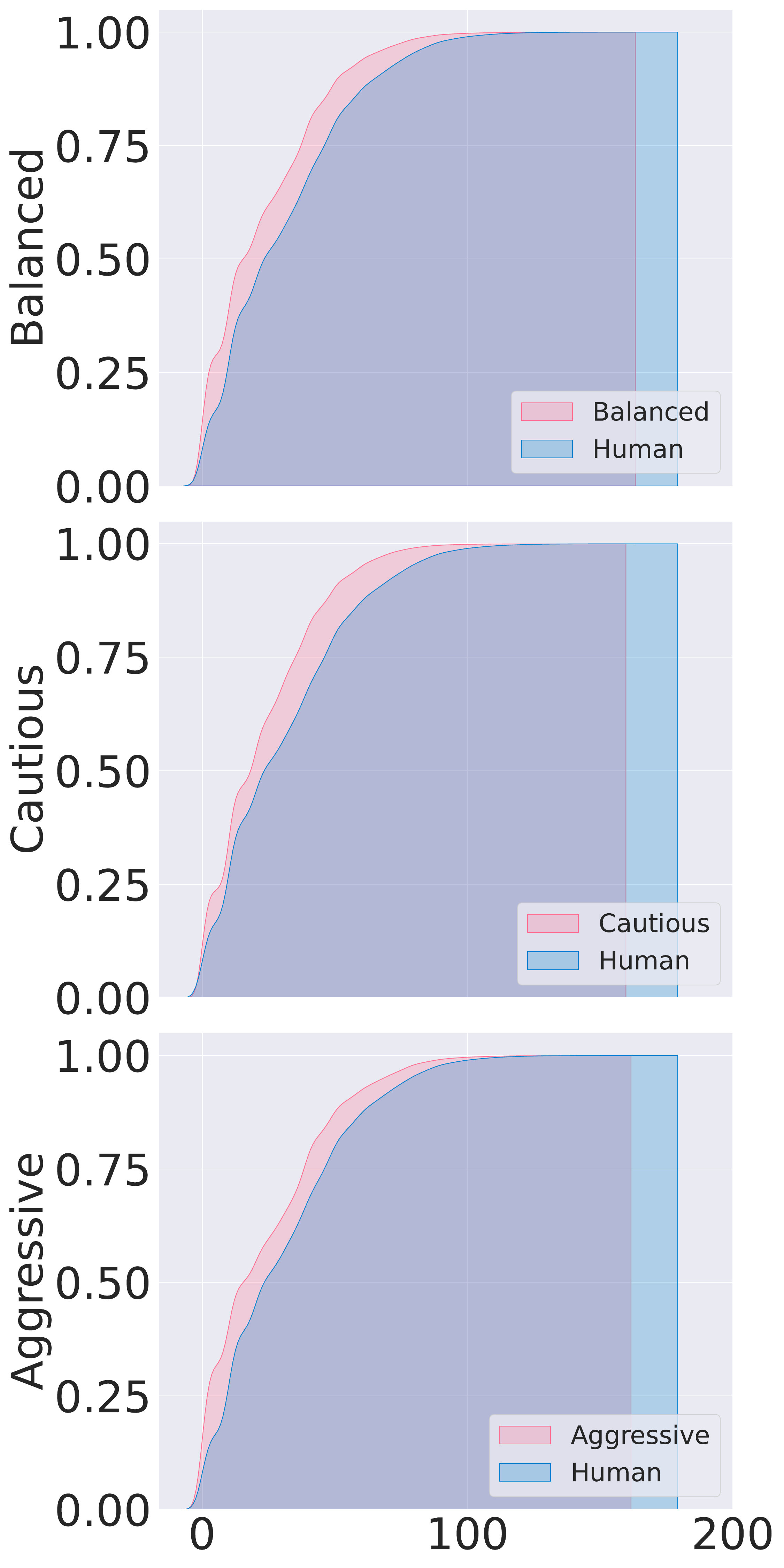}\label{Blitz}} \subfigure[Counter] {\includegraphics[width=.18\textwidth]{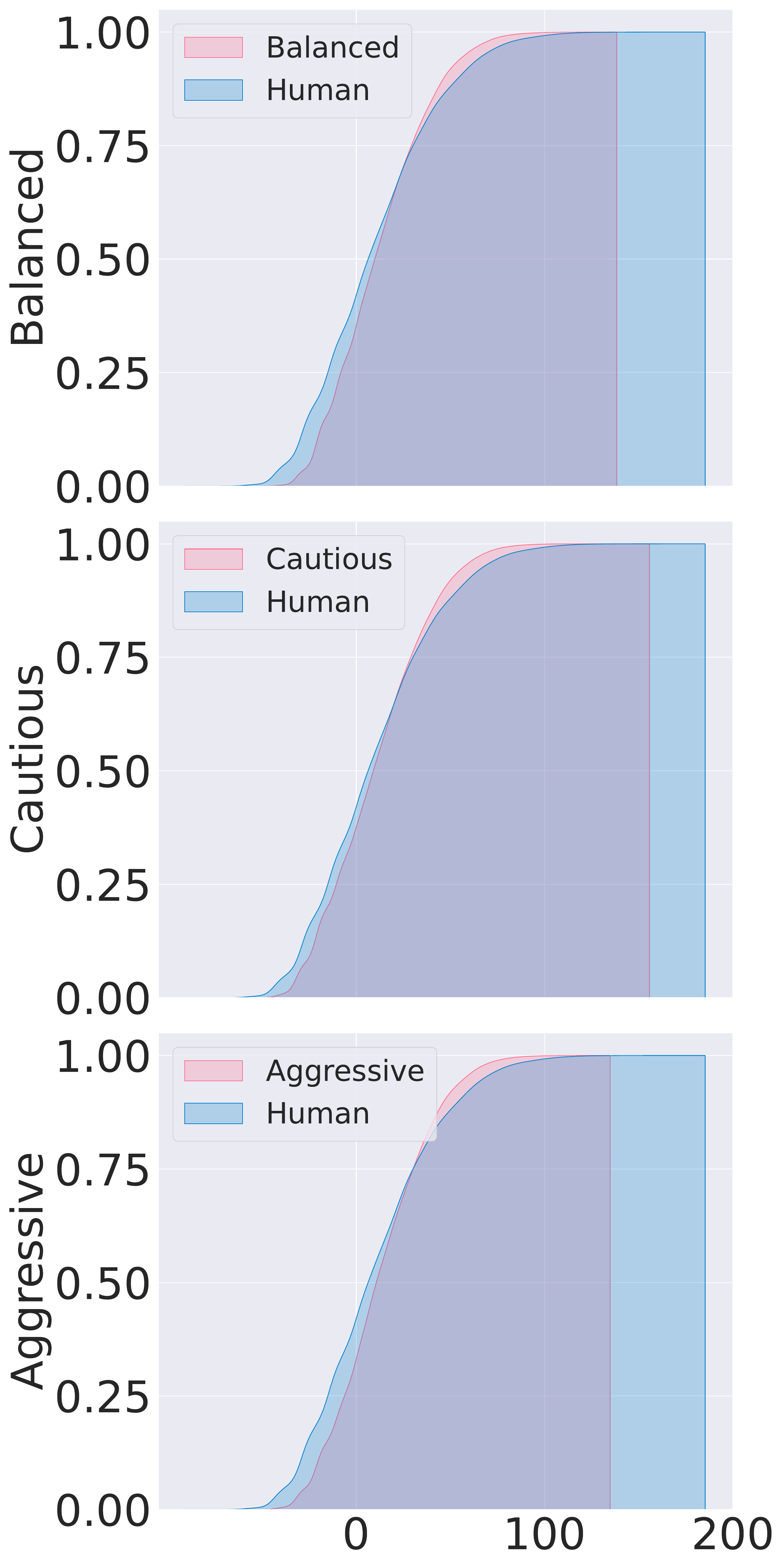}\label{counter}} \subfigure[Attack] {\includegraphics[width=.18\textwidth]{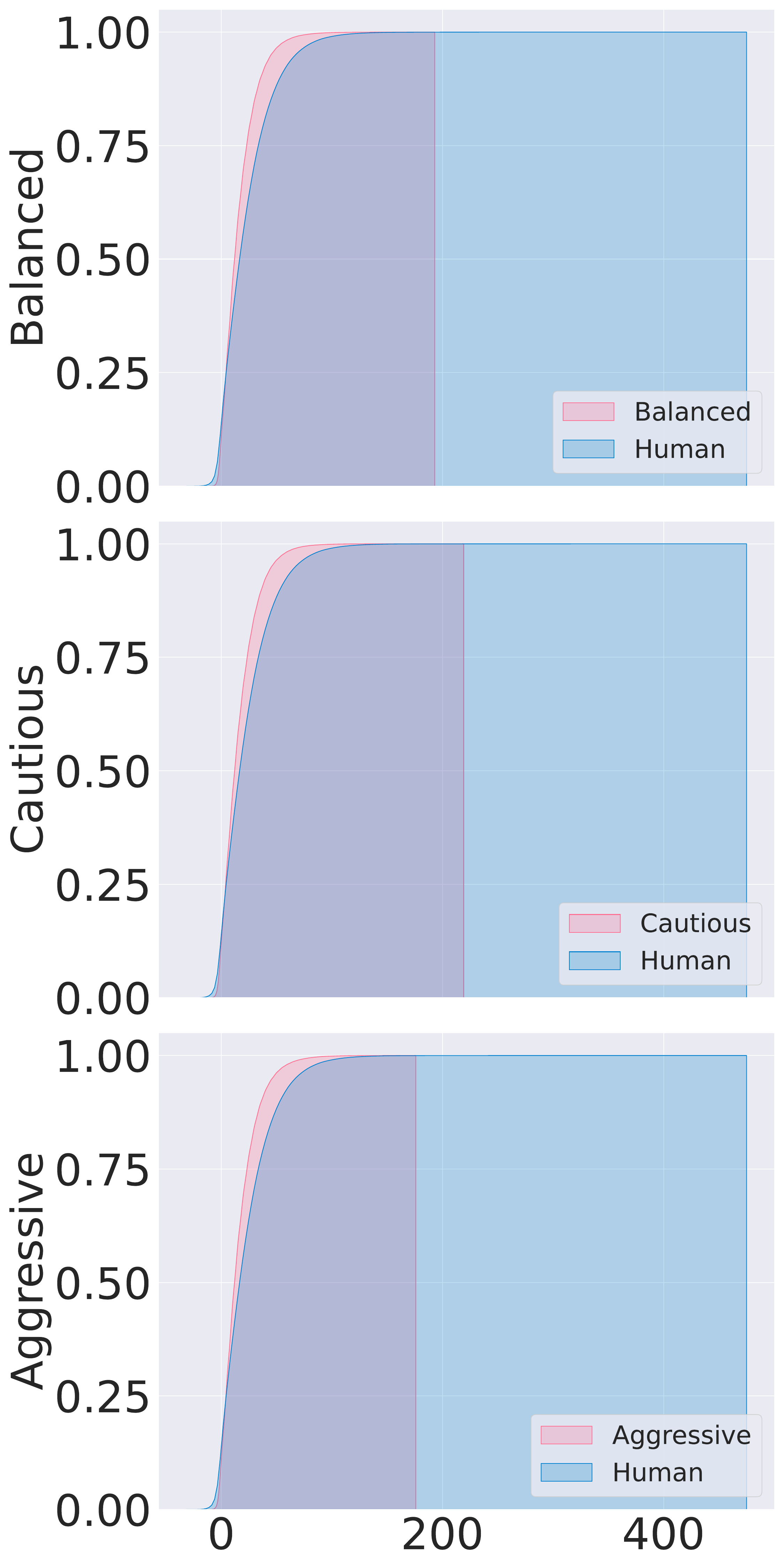}\label{Attack}} 
\caption{A comparison of the cumulative distribution function (CDF) between human players and different agents across different metrics. The x-axis represents the reward scores, while the y-axis represents the probabilities. Higher reward scores indicate better behaviors. Agents that were induced with behavior through three types of rewards mentioned in \cref{methods_align}, over 120,000 game instances have been collected for comparison with human players.} 
\label{fig_distri} 
\vskip 0.1in
\end{figure*}

\subsection{Agent-Human Alignment}
To discuss the consistency between DRL agents and human players, we have designed a scoring system for detecting in-game behaviors based on human expert priors. This scoring system can score based on the semantic-level behavior of human players or DRL agents in the game, such as the rationality of using skills in a certain situation, etc. \cref{fig_distri} displays the differences in cumulative distribution function (CDF) between human players and different DRL agents across Five metrics. The x-axis represents the reward scores, while the y-axis represents the probabilities. Higher reward scores indicate better behaviors.

\textbf{Metrics.} \textbf{Substitution} indicates the release of ultimate defensive skills, \textbf{Special} represents the release of the highest damage-dealing skill, \textbf{Blitz} indicates a preference for taking the initiative at the beginning of the game, \textbf{Counter} indicates a tendency to counter-attack at the beginning of the game, and \textbf{Attack} represents regular attacks without using skills. Agents that were induced with behavior through three types of rewards mentioned in \cref{methods_align}, over 120,000 game instances have been collected for comparison with human players. These players, with Elo scores of 1500 or above, exhibit their experience and expertise in the Naruto Mobile game. 

The result in \cref{counter} indicates that DRL agents are generally more cautious than human players. However, different reward functions still have an impact on the behavior of the DRL agents. The aggressive agent receives lower rewards in the \textbf{Counter} metric compared to the cautious agent, while the balanced agent falls in between the two. The results in \cref{Substitution} describes \textit{Shūkai} are more precise in releasing defensive skills compared to human players, which is why they consistently outperform in \cref{counter}. In \cref{Attack}, \textit{Shūkai} do not prioritize regular attacks. This can be attributed to the \textit{Shūkai}'s precise skill releases (\cref{Special}).

In conclusion, \textit{Shūkai} exhibit more precise skill releases but tend to be more cautious than humans. However, this also demonstrates that we can induce different behavioral styles in the agents through special reward functions, which can guide us in designing agents with specific styles. The evaluation system works to effectively observe the gap between human players and DRL agents, helping design more human-like agents to enhance the gaming experience. Overall, \textit{Shūkai} still has some gap compared to human players in terms of high reward scores, indicating the need for further improvement in our algorithms to encourage more advanced exploration by the agents. Additional detailed data can be found in the \cref{a_human_align}. 
\section{Related Work}
\subsection{DRL for Games}
With the development of DRL, there are several successful applications in different domains such as robotic \citep{andrychowicz2017hindsight}, autonomous vehicles \citep{aradi2020survey}, and video games. 
DQN \citep{mnih2015human} proposed trying its hand at Atari games using DRL from pixel input. AlphaGo \citep{silver2016mastering, silver2017mastering,silver2018general} solved the game of Go with self-play. Researchers have discovered the ability of self-play to solve game problems, which has sparked interest among researchers to apply it in the field of game theory. applied similar methods to card games. \citep{moravvcik2017deepstack,zha2021douzero, zhao2022douzero+} applied similar methods to card games such as Texas Hold'em and DouDizhu (a popular Chinese card game). AlphaStar \citep{vinyals2019grandmaster} and OpenAI Five \citep{berner2019dota} used reinforcement learning to train game AI in two famous games, StarCraft II and Dota 2. \citet{jaderberg2019human} achieved human performance in Capture the Flag.
Subsequently, the AI they designed in the MOBA game King of Glory demonstrated outstanding performance. Subsequently, the AI designed by \cite{ye2020mastering} and \citep{ye2020towards} in the MOBA game King of Glory demonstrated outstanding performance. Additional discussion can be found in the \cref{a_Related}.

\subsection{DRL for Fighting Games}
In the field of game AI for fighting games, there are numerous notable precedents, primarily focused on the application scenario of FightICE\citep{khan2022darefightingice}. \citet{kim2018hybrid} combined genetic algorithms and Monte Carlo Tree Search (MCTS) to find general solutions that address the constraints of game AI while maintaining competitiveness. \citet{ishii2018monte} introduced the concept of aligning with user personas in game AI, which has gained significant attention in recent research. \citet{kim2020mastering} introduced a reinforcement learning approach combined with MCTS to enhance the performance of the game AI and proposed novel evaluation metrics to ensure the stability of game AI. \citet{oh2021creating} introduced self-play curriculum learning and three types of rewards to further enhance AI personas and performance. \citet{halina2022diversity} introduced a diversity-based deep reinforcement learning approach for generating a set of agents with similar difficulty but utilizing diverse strategies. \citet{ishii2018monte} and \citet{halina2022diversity} share similarities with our research, but the core attention of HELT lies in addressing the training efficiency and generalization issues caused by a large pool of characters. Additionally, we introduce multi-style rewards to align with players and ensure a satisfactory gaming experience.
\section{Conclusions}

In this paper, we propose a practical DRL agent system called \textit{Shūkai} to address key challenges in its deployment to the fighting game Naruto Mobile, focusing on achieving a balance between training efficiency, model competence, and generalizability. This is achieved by utilizing a unified model to control multiple characters simultaneously, quantifying input states to enhance generalization, and introducing HELT to accelerate training efficiency. Experimental results demonstrate the powerful capabilities of \textit{Shūkai}, which has been adopted by Naruto Mobile since 2022 for player training, aiding in skill improvement. Furthermore, we present the concept of aligning DRL agent behavior with human players, analyzing the differences between DRL agents and human players through reward and practical evaluations. This is crucial for the practical application of DRL agents in commercial games. With the integration of \textit{Shūkai}, the game has witnessed a notable 5\% maximum increase in the retention rate and an average improvement of 4\%.
\section{Future Work}

\textit{Shūkai} has been employed by Naruto Mobile to train its agents and have achieved great results. Our future work lies in extending the principles of \textit{Shūkai} and applying them to more games and domains. We have successfully deployed \textit{Shūkai} to two new fighting games that are currently under development at Tencent. Additionally, \textit{Shūkai} is also planned to be used in Arena Breakout, a first-person shooter game published by Tencent. 

We will continue to promote the practical implementation of reinforcement learning and further its theoretical development, enabling reinforcement learning to better empower the real world.
\section*{Acknowledgement}
This project was partially supported by the National Natural Science Foundation of China (No.62072427, No.12227901), the Project of Stable Support for Youth Team in Basic Research Field, CAS (No.YSBR-005), Academic Leaders Cultivation Program, USTC.
\section*{Impact Statement}
This paper presents work that aims to advance the field of Machine Learning. This work marks a further step in the deployment of DRL systems in a large-scale commercial fighting game. To the best of our knowledge, this is the first example of a DRL system being employed by a commercial fighting game.

Reinforcement learning has seen applications and advancements in various domains as a means to solve sequential decision-making problems. We aim to promote the coexistence of better reinforcement learning methods with human applications. In the Introduction section, we mentioned that most previous work focused on training stronger and more competitive agents. However, as game players, we don't necessarily want to compete against unbeatable agents. It is based on this understanding that we introduced \textit{Shūkai}. To address the training and generalization challenges posed by a large character pool, we proposed HELT and conducted initial explorations of human-agent alignment. It's important to emphasize that this is a practical implementation of reinforcement learning. While our focus was on Naruto Mobile in this paper, the HELT approach and the concept of human-agent alignment can be applied to reinforcement learning in general, regardless of whether it's within a game or in other contexts.

Returning to games, having controllable anthropomorphic agents in the game ecosystem benefits both players and game developers. It was observed that human players are enthusiastic about immersing themselves in the story and becoming part of it. This means that highly anthropomorphic agents can help players derive more enjoyment from their gaming experience, thereby promoting player engagement and social exploration within the game. This represents a positive influence. For game developers, having controllable anthropomorphic agents can contribute to the creation of better activities and gameplay designs, ultimately producing superior games.

\nocite{langley00}

\bibliography{example_paper}

\begin{thebibliography}{60}
\providecommand{\natexlab}[1]{#1}
\providecommand{\url}[1]{\texttt{#1}}
\expandafter\ifx\csname urlstyle\endcsname\relax
  \providecommand{\doi}[1]{doi: #1}\else
  \providecommand{\doi}{doi: \begingroup \urlstyle{rm}\Url}\fi

\bibitem[Andrychowicz et~al.(2017)Andrychowicz, Wolski, Ray, Schneider, Fong, Welinder, McGrew, Tobin, Pieter~Abbeel, and Zaremba]{andrychowicz2017hindsight}
Andrychowicz, M., Wolski, F., Ray, A., Schneider, J., Fong, R., Welinder, P., McGrew, B., Tobin, J., Pieter~Abbeel, O., and Zaremba, W.
\newblock Hindsight experience replay.
\newblock \emph{Advances in Neural Information Processing Systems (NeurIPS)}, 30, 2017.

\bibitem[Aradi(2020)]{aradi2020survey}
Aradi, S.
\newblock Survey of deep reinforcement learning for motion planning of autonomous vehicles.
\newblock \emph{IEEE Transactions on Intelligent Transportation Systems}, 23\penalty0 (2):\penalty0 740--759, 2020.

\bibitem[Bakker et~al.(2011)Bakker, Oerlemans, Demerouti, Slot, and Ali]{bakker2011flow}
Bakker, A.~B., Oerlemans, W., Demerouti, E., Slot, B.~B., and Ali, D.~K.
\newblock Flow and performance: A study among talented dutch soccer players.
\newblock \emph{Psychology of Sport and Exercise}, 12\penalty0 (4):\penalty0 442--450, 2011.

\bibitem[Bejani \& Ghatee(2021)Bejani and Ghatee]{bejani2021systematic}
Bejani, M.~M. and Ghatee, M.
\newblock A systematic review on overfitting control in shallow and deep neural networks.
\newblock \emph{Artificial Intelligence Review}, pp.\  1--48, 2021.

\bibitem[Berner et~al.(2019)Berner, Brockman, Chan, Cheung, Debiak, Dennison, Farhi, Fischer, Hashme, Hesse, et~al.]{berner2019dota}
Berner, C., Brockman, G., Chan, B., Cheung, V., Debiak, P., Dennison, C., Farhi, D., Fischer, Q., Hashme, S., Hesse, C., et~al.
\newblock Dota 2 with large scale deep reinforcement learning.
\newblock \emph{arXiv preprint arXiv:1912.06680}, 2019.

\bibitem[Brown(1951)]{brown1951iterative}
Brown, G.~W.
\newblock Iterative solution of games by fictitious play.
\newblock \emph{Act. Anal. Prod Allocation}, 13\penalty0 (1):\penalty0 374, 1951.

\bibitem[Brown \& Sandholm(2018)Brown and Sandholm]{brown2018superhuman}
Brown, N. and Sandholm, T.
\newblock Superhuman ai for heads-up no-limit poker: Libratus beats top professionals.
\newblock \emph{Science}, 359\penalty0 (6374):\penalty0 418--424, 2018.

\bibitem[Cheng et~al.(2016)Cheng, Hung, and Chen]{cheng2016influence}
Cheng, T.-M., Hung, S.-H., and Chen, M.-T.
\newblock The influence of leisure involvement on flow experience during hiking activity: Using psychological commitment as a mediate variable.
\newblock \emph{Asia Pacific Journal of Tourism Research}, 21\penalty0 (1):\penalty0 1--19, 2016.

\bibitem[Cobbe et~al.(2019)Cobbe, Klimov, Hesse, Kim, and Schulman]{cobbe2019quantifying}
Cobbe, K., Klimov, O., Hesse, C., Kim, T., and Schulman, J.
\newblock Quantifying generalization in reinforcement learning.
\newblock In \emph{International Conference on Machine Learning (ICML)}, pp.\  1282--1289. PMLR, 2019.

\bibitem[Csikszentmihalyi(2000)]{csikszentmihalyi2000beyond}
Csikszentmihalyi, M.
\newblock \emph{Beyond boredom and anxiety.}
\newblock Jossey-bass, 2000.

\bibitem[Dosovitskiy \& Koltun(2016)Dosovitskiy and Koltun]{dosovitskiy2016learning}
Dosovitskiy, A. and Koltun, V.
\newblock Learning to act by predicting the future.
\newblock \emph{arXiv preprint arXiv:1611.01779}, 2016.

\bibitem[Elo(1978)]{elo1978rating}
Elo, A.
\newblock The rating of chess players, past and present (arco, new york).
\newblock 1978.

\bibitem[Finn et~al.(2017)Finn, Abbeel, and Levine]{finn2017model}
Finn, C., Abbeel, P., and Levine, S.
\newblock Model-agnostic meta-learning for fast adaptation of deep networks.
\newblock In \emph{International Conference On Machine Learning (ICML)}, pp.\  1126--1135. PMLR, 2017.

\bibitem[Haarnoja et~al.(2018)Haarnoja, Zhou, Abbeel, and Levine]{sac}
Haarnoja, T., Zhou, A., Abbeel, P., and Levine, S.
\newblock Soft actor-critic: Off-policy maximum entropy deep reinforcement learning with a stochastic actor.
\newblock In \emph{International Conference on Machine Learning (ICML)}, pp.\  1861--1870, 2018.

\bibitem[Hafner et~al.(2019)Hafner, Lillicrap, Fischer, Villegas, Ha, Lee, and Davidson]{hafner2019learning}
Hafner, D., Lillicrap, T.~P., Fischer, I., Villegas, R., Ha, D., Lee, H., and Davidson, J.
\newblock Learning latent dynamics for planning from pixels.
\newblock In \emph{Proceedings of the 36th International Conference on Machine Learning}, volume~97, pp.\  2467--2475, 2019.

\bibitem[Halina \& Guzdial(2022)Halina and Guzdial]{halina2022diversity}
Halina, E. and Guzdial, M.
\newblock Diversity-based deep reinforcement learning towards multidimensional difficulty for fighting game ai.
\newblock \emph{arXiv preprint arXiv:2211.02759}, 2022.

\bibitem[He et~al.(2023)He, Su, Zhang, and Hou]{DBLP:conf/cvpr/HeSZH23}
He, Q., Su, H., Zhang, J., and Hou, X.
\newblock Frustratingly easy regularization on representation can boost deep reinforcement learning.
\newblock In \emph{{IEEE/CVF} Conference on Computer Vision and Pattern Recognition, {CVPR} 2023, Vancouver, BC, Canada, June 17-24, 2023}, pp.\  20215--20225. {IEEE}, 2023.
\newblock \doi{10.1109/CVPR52729.2023.01936}.
\newblock URL \url{https://doi.org/10.1109/CVPR52729.2023.01936}.

\bibitem[He et~al.(2024)He, Zhou, Fang, and Maghsudi]{DBLP:journals/corr/abs-2404-12754}
He, Q., Zhou, T., Fang, M., and Maghsudi, S.
\newblock Adaptive regularization of representation rank as an implicit constraint of bellman equation.
\newblock \emph{CoRR}, abs/2404.12754, 2024.
\newblock \doi{10.48550/ARXIV.2404.12754}.
\newblock URL \url{https://doi.org/10.48550/arXiv.2404.12754}.

\bibitem[Heinrich \& Silver(2016)Heinrich and Silver]{heinrich2016deep}
Heinrich, J. and Silver, D.
\newblock Deep reinforcement learning from self-play in imperfect-information games, 2016.

\bibitem[Heinrich et~al.(2015)Heinrich, Lanctot, and Silver]{Heinrich2015FictitiousSI}
Heinrich, J., Lanctot, M., and Silver, D.
\newblock Fictitious self-play in extensive-form games.
\newblock In \emph{International Conference on Machine Learning (ICML)}, 2015.

\bibitem[Houthooft et~al.(2016)Houthooft, Chen, Isola, Cruz, and Abbeel]{houthooft2016vime}
Houthooft, R., Chen, X., Isola, P., Cruz, E., and Abbeel, P.
\newblock Vime: Variational information maximizing exploration.
\newblock In \emph{Advances in Neural Information Processing Systems}, pp.\  1109--1117, 2016.

\bibitem[Ishii et~al.(2018)Ishii, Ito, Ishihara, Harada, and Thawonmas]{ishii2018monte}
Ishii, R., Ito, S., Ishihara, M., Harada, T., and Thawonmas, R.
\newblock Monte-carlo tree search implementation of fighting game ais having personas.
\newblock In \emph{2018 IEEE Conference on Computational Intelligence and Games (CIG)}, pp.\  1--8. IEEE, 2018.

\bibitem[Jaderberg et~al.(2019)Jaderberg, Czarnecki, Dunning, Marris, Lever, Castaneda, Beattie, Rabinowitz, Morcos, Ruderman, et~al.]{jaderberg2019human}
Jaderberg, M., Czarnecki, W.~M., Dunning, I., Marris, L., Lever, G., Castaneda, A.~G., Beattie, C., Rabinowitz, N.~C., Morcos, A.~S., Ruderman, A., et~al.
\newblock Human-level performance in 3d multiplayer games with population-based reinforcement learning.
\newblock \emph{Science}, 364\penalty0 (6443):\penalty0 859--865, 2019.

\bibitem[Khan et~al.(2022)Khan, Van~Nguyen, Dai, and Thawonmas]{khan2022darefightingice}
Khan, I., Van~Nguyen, T., Dai, X., and Thawonmas, R.
\newblock Darefightingice competition: A fighting game sound design and ai competition.
\newblock In \emph{2022 IEEE Conference on Games (CoG)}, pp.\  478--485. IEEE, 2022.

\bibitem[Kim et~al.(2020)Kim, Park, and Yang]{kim2020mastering}
Kim, D.-W., Park, S., and Yang, S.-i.
\newblock Mastering fighting game using deep reinforcement learning with self-play.
\newblock In \emph{2020 IEEE Conference on Games (CoG)}, pp.\  576--583. IEEE, 2020.

\bibitem[Kim \& Ahn(2018)Kim and Ahn]{kim2018hybrid}
Kim, M.-J. and Ahn, C.~W.
\newblock Hybrid fighting game ai using a genetic algorithm and monte carlo tree search.
\newblock In \emph{Proceedings of the Genetic And Evolutionary Computation Conference Companion}, pp.\  129--130, 2018.

\bibitem[Lan et~al.(2024)Lan, Zhang, Yi, Guo, Peng, Gao, Wu, Chen, Du, Hu, et~al.]{lan2024contrastive}
Lan, S., Zhang, R., Yi, Q., Guo, J., Peng, S., Gao, Y., Wu, F., Chen, R., Du, Z., Hu, X., et~al.
\newblock Contrastive modules with temporal attention for multi-task reinforcement learning.
\newblock \emph{Advances in Neural Information Processing Systems}, 36, 2024.

\bibitem[Lanctot et~al.(2017)Lanctot, Zambaldi, Gruslys, Lazaridou, Tuyls, P{\'e}rolat, Silver, and Graepel]{lanctot2017unified}
Lanctot, M., Zambaldi, V., Gruslys, A., Lazaridou, A., Tuyls, K., P{\'e}rolat, J., Silver, D., and Graepel, T.
\newblock A unified game-theoretic approach to multiagent reinforcement learning.
\newblock \emph{Advances in Neural Information Processing Systems (NeurIPS)}, 30, 2017.

\bibitem[Leslie \& Collins(2006)Leslie and Collins]{leslie2006generalised}
Leslie, D.~S. and Collins, E.~J.
\newblock Generalised weakened fictitious play.
\newblock \emph{Games and Economic Behavior}, 56\penalty0 (2):\penalty0 285--298, 2006.

\bibitem[Li et~al.(2023)Li, Lyu, Ma, Wang, Yang, Li, and Li]{li2023normalization}
Li, L., Lyu, J., Ma, G., Wang, Z., Yang, Z., Li, X., and Li, Z.
\newblock Normalization enhances generalization in visual reinforcement learning.
\newblock \emph{arXiv preprint arXiv:2306.00656}, 2023.

\bibitem[Lyu et~al.(2024{\natexlab{a}})Lyu, Bai, Yang, Lu, and Li]{lyu2024cross}
Lyu, J., Bai, C., Yang, J., Lu, Z., and Li, X.
\newblock Cross-domain policy adaptation by capturing representation mismatch.
\newblock \emph{arXiv preprint arXiv:2405.15369}, 2024{\natexlab{a}}.

\bibitem[Lyu et~al.(2024{\natexlab{b}})Lyu, Wan, Li, and Lu]{lyu2024towards}
Lyu, J., Wan, L., Li, X., and Lu, Z.
\newblock Towards understanding how to reduce generalization gap in visual reinforcement learning.
\newblock In \emph{Proceedings of the 23rd International Conference on Autonomous Agents and Multiagent Systems}, pp.\  2369--2371, 2024{\natexlab{b}}.

\bibitem[Lyu et~al.(2024{\natexlab{c}})Lyu, Wan, Li, and Lu]{lyu2024understanding}
Lyu, J., Wan, L., Li, X., and Lu, Z.
\newblock Understanding what affects generalization gap in visual reinforcement learning: Theory and empirical evidence.
\newblock \emph{arXiv preprint arXiv:2402.02701}, 2024{\natexlab{c}}.

\bibitem[McMahan et~al.(2003)McMahan, Gordon, and Blum]{mcmahan2003planning}
McMahan, H.~B., Gordon, G.~J., and Blum, A.
\newblock Planning in the presence of cost functions controlled by an adversary.
\newblock In \emph{International Conference on Machine Learning (ICML)}, pp.\  536--543, 2003.

\bibitem[Mnih et~al.(2015)Mnih, Kavukcuoglu, Silver, Rusu, Veness, Bellemare, Graves, Riedmiller, Fidjeland, Ostrovski, et~al.]{mnih2015human}
Mnih, V., Kavukcuoglu, K., Silver, D., Rusu, A.~A., Veness, J., Bellemare, M.~G., Graves, A., Riedmiller, M., Fidjeland, A.~K., Ostrovski, G., et~al.
\newblock Human-level control through deep reinforcement learning.
\newblock \emph{Nature}, 518\penalty0 (7540):\penalty0 529--533, 2015.

\bibitem[Morav{\v{c}}{\'\i}k et~al.(2017)Morav{\v{c}}{\'\i}k, Schmid, Burch, Lis{\`y}, Morrill, Bard, Davis, Waugh, Johanson, and Bowling]{moravvcik2017deepstack}
Morav{\v{c}}{\'\i}k, M., Schmid, M., Burch, N., Lis{\`y}, V., Morrill, D., Bard, N., Davis, T., Waugh, K., Johanson, M., and Bowling, M.
\newblock Deepstack: Expert-level artificial intelligence in heads-up no-limit poker.
\newblock \emph{Science}, 356\penalty0 (6337):\penalty0 508--513, 2017.

\bibitem[Oh et~al.(2021)Oh, Rho, Moon, Son, Lee, and Chung]{oh2021creating}
Oh, I., Rho, S., Moon, S., Son, S., Lee, H., and Chung, J.
\newblock Creating pro-level ai for a real-time fighting game using deep reinforcement learning.
\newblock \emph{IEEE Transactions on Games}, 14\penalty0 (2):\penalty0 212--220, 2021.

\bibitem[Ouyang et~al.(2022)Ouyang, Wu, Jiang, Almeida, Wainwright, Mishkin, Zhang, Agarwal, Slama, Ray, et~al.]{instructgpt}
Ouyang, L., Wu, J., Jiang, X., Almeida, D., Wainwright, C., Mishkin, P., Zhang, C., Agarwal, S., Slama, K., Ray, A., et~al.
\newblock Training language models to follow instructions with human feedback.
\newblock \emph{Advances in Neural Information Processing Systems (NeurIPS)}, 35:\penalty0 27730--27744, 2022.

\bibitem[Pathak et~al.(2017)Pathak, Agrawal, Efros, and Darrell]{pathak2017curiosity}
Pathak, D., Agrawal, P., Efros, A.~A., and Darrell, T.
\newblock Curiosity-driven exploration by self-supervised prediction.
\newblock In \emph{Proceedings of the IEEE Conference on Computer Vision and Pattern Recognition}, pp.\  16--17, 2017.

\bibitem[Peng et~al.(2023{\natexlab{a}})Peng, Hu, Yi, Zhang, Guo, Huang, Tian, Chen, Du, Guo, Chen, and Li]{Peng2023SelfdrivenGL}
Peng, S., Hu, X., Yi, Q., Zhang, R., Guo, J., Huang, D., Tian, Z., Chen, R., Du, Z., Guo, Q., Chen, Y., and Li, L.
\newblock Self-driven grounding: Large language model agents with automatical language-aligned skill learning.
\newblock \emph{ArXiv}, abs/2309.01352, 2023{\natexlab{a}}.
\newblock URL \url{https://api.semanticscholar.org/CorpusID:261530737}.

\bibitem[Peng et~al.(2023{\natexlab{b}})Peng, Hu, Zhang, Guo, Yi, Chen, Du, Li, Guo, and Chen]{Peng2023ConceptualRL}
Peng, S., Hu, X., Zhang, R., Guo, J., Yi, Q., Chen, R., Du, Z., Li, L., Guo, Q., and Chen, Y.
\newblock Conceptual reinforcement learning for language-conditioned tasks.
\newblock In \emph{AAAI Conference on Artificial Intelligence}, 2023{\natexlab{b}}.
\newblock URL \url{https://api.semanticscholar.org/CorpusID:257427058}.

\bibitem[Schaul et~al.(2015)Schaul, Horgan, Gregor, and Silver]{schaul2015universal}
Schaul, T., Horgan, D., Gregor, K., and Silver, D.
\newblock Universal value function approximators.
\newblock In \emph{International Conference On Machine Learning (ICML)}, pp.\  1312--1320. PMLR, 2015.

\bibitem[Schulman et~al.(2015)Schulman, Moritz, Levine, Jordan, and Abbeel]{schulman2015high}
Schulman, J., Moritz, P., Levine, S., Jordan, M., and Abbeel, P.
\newblock High-dimensional continuous control using generalized advantage estimation.
\newblock \emph{arXiv preprint arXiv:1506.02438}, 2015.

\bibitem[Schulman et~al.(2017)Schulman, Wolski, Dhariwal, Radford, and Klimov]{schulman2017proximal}
Schulman, J., Wolski, F., Dhariwal, P., Radford, A., and Klimov, O.
\newblock Proximal policy optimization algorithms, 2017.

\bibitem[Shafer(2012)]{shafer2012causes}
Shafer, D.~M.
\newblock Causes of state hostility and enjoyment in player versus player and player versus environment video games.
\newblock \emph{Journal of Communication}, 62\penalty0 (4):\penalty0 719--737, 2012.

\bibitem[Silver et~al.(2016)Silver, Huang, Maddison, Guez, Sifre, Van Den~Driessche, Schrittwieser, Antonoglou, Panneershelvam, Lanctot, et~al.]{silver2016mastering}
Silver, D., Huang, A., Maddison, C.~J., Guez, A., Sifre, L., Van Den~Driessche, G., Schrittwieser, J., Antonoglou, I., Panneershelvam, V., Lanctot, M., et~al.
\newblock Mastering the game of go with deep neural networks and tree search.
\newblock \emph{Nature}, 529\penalty0 (7587):\penalty0 484--489, 2016.

\bibitem[Silver et~al.(2017)Silver, Schrittwieser, Simonyan, Antonoglou, Huang, Guez, Hubert, Baker, Lai, Bolton, et~al.]{silver2017mastering}
Silver, D., Schrittwieser, J., Simonyan, K., Antonoglou, I., Huang, A., Guez, A., Hubert, T., Baker, L., Lai, M., Bolton, A., et~al.
\newblock Mastering the game of go without human knowledge.
\newblock \emph{Nature}, 550\penalty0 (7676):\penalty0 354--359, 2017.

\bibitem[Silver et~al.(2018)Silver, Hubert, Schrittwieser, Antonoglou, Lai, Guez, Lanctot, Sifre, Kumaran, Graepel, et~al.]{silver2018general}
Silver, D., Hubert, T., Schrittwieser, J., Antonoglou, I., Lai, M., Guez, A., Lanctot, M., Sifre, L., Kumaran, D., Graepel, T., et~al.
\newblock A general reinforcement learning algorithm that masters chess, shogi, and go through self-play.
\newblock \emph{Science}, 362\penalty0 (6419):\penalty0 1140--1144, 2018.

\bibitem[Sun et~al.(2017)Sun, Hoffman, Saenko, and Darrell]{sun2017learning}
Sun, X., Hoffman, J., Saenko, K., and Darrell, T.
\newblock Learning to learn how to learn: Self-adaptive visual navigation using meta-learning.
\newblock In \emph{Proceedings of the IEEE International Conference on Computer Vision}, pp.\  675--684, 2017.

\bibitem[Sutton \& Barto(2018)Sutton and Barto]{sutton2018reinforcement}
Sutton, R.~S. and Barto, A.~G.
\newblock \emph{Reinforcement learning: An introduction}.
\newblock MIT press, 2018.

\bibitem[Touvron et~al.(2023)Touvron, Martin, Stone, Albert, Almahairi, Babaei, Bashlykov, Batra, Bhargava, Bhosale, et~al.]{llama2}
Touvron, H., Martin, L., Stone, K., Albert, P., Almahairi, A., Babaei, Y., Bashlykov, N., Batra, S., Bhargava, P., Bhosale, S., et~al.
\newblock Llama 2: Open foundation and fine-tuned chat models.
\newblock \emph{arXiv preprint arXiv:2307.09288}, 2023.

\bibitem[Vinyals et~al.(2019)Vinyals, Babuschkin, Czarnecki, Mathieu, Dudzik, Chung, Choi, Powell, Ewalds, Georgiev, et~al.]{vinyals2019grandmaster}
Vinyals, O., Babuschkin, I., Czarnecki, W.~M., Mathieu, M., Dudzik, A., Chung, J., Choi, D.~H., Powell, R., Ewalds, T., Georgiev, P., et~al.
\newblock Grandmaster level in starcraft ii using multi-agent reinforcement learning.
\newblock \emph{Nature}, 575\penalty0 (7782):\penalty0 350--354, 2019.

\bibitem[Wang et~al.(2023)Wang, Liang, Li, Li, Ghanem, Zimmermann, Yi, Zhang, Wang, et~al.]{wang2023brave}
Wang, K., Liang, Y., Li, X., Li, G., Ghanem, B., Zimmermann, R., Yi, H., Zhang, Y., Wang, Y., et~al.
\newblock Brave the wind and the waves: Discovering robust and generalizable graph lottery tickets.
\newblock \emph{IEEE Transactions on Pattern Analysis and Machine Intelligence}, 2023.

\bibitem[Wang(2024)]{wang2024balancing}
Wang, X.
\newblock Balancing the ai strength of roles in self-play training with regret matching+, 2024.

\bibitem[Ye et~al.(2020{\natexlab{a}})Ye, Chen, Zhang, Chen, Yuan, Liu, Chen, Liu, Qiu, Yu, et~al.]{ye2020towards}
Ye, D., Chen, G., Zhang, W., Chen, S., Yuan, B., Liu, B., Chen, J., Liu, Z., Qiu, F., Yu, H., et~al.
\newblock Towards playing full moba games with deep reinforcement learning.
\newblock \emph{Advances in Neural Information Processing Systems (NeurIPS)}, 33:\penalty0 621--632, 2020{\natexlab{a}}.

\bibitem[Ye et~al.(2020{\natexlab{b}})Ye, Liu, Sun, Shi, Zhao, Wu, Yu, Yang, Wu, Guo, et~al.]{ye2020mastering}
Ye, D., Liu, Z., Sun, M., Shi, B., Zhao, P., Wu, H., Yu, H., Yang, S., Wu, X., Guo, Q., et~al.
\newblock Mastering complex control in moba games with deep reinforcement learning.
\newblock In \emph{AAAI Conference on Artificial Intelligence (AAAI)}, volume~34, pp.\  6672--6679, 2020{\natexlab{b}}.

\bibitem[Yi et~al.(2022)Yi, Zhang, Peng, Guo, Hu, Du, Zhang, Guo, and Chen]{DBLP:conf/nips/Yi0PG0Dz0C22}
Yi, Q., Zhang, R., Peng, S., Guo, J., Hu, X., Du, Z., Zhang, X., Guo, Q., and Chen, Y.
\newblock Object-category aware reinforcement learning.
\newblock In Koyejo, S., Mohamed, S., Agarwal, A., Belgrave, D., Cho, K., and Oh, A. (eds.), \emph{Advances in Neural Information Processing Systems 35: Annual Conference on Neural Information Processing Systems 2022, NeurIPS 2022, New Orleans, LA, USA, November 28 - December 9, 2022}, 2022.

\bibitem[Yi et~al.(2023)Yi, Zhang, Peng, Guo, Gao, Yuan, Chen, Lan, Hu, Du, Zhang, Guo, and Chen]{DBLP:conf/icml/Yi0PGGYCLHDZGC23}
Yi, Q., Zhang, R., Peng, S., Guo, J., Gao, Y., Yuan, K., Chen, R., Lan, S., Hu, X., Du, Z., Zhang, X., Guo, Q., and Chen, Y.
\newblock Online prototype alignment for few-shot policy transfer.
\newblock In Krause, A., Brunskill, E., Cho, K., Engelhardt, B., Sabato, S., and Scarlett, J. (eds.), \emph{International Conference on Machine Learning, {ICML} 2023, 23-29 July 2023, Honolulu, Hawaii, {USA}}, volume 202 of \emph{Proceedings of Machine Learning Research}, pp.\  39968--39983. {PMLR}, 2023.
\newblock URL \url{https://proceedings.mlr.press/v202/yi23b.html}.

\bibitem[Zha et~al.(2021)Zha, Xie, Ma, Zhang, Lian, Hu, and Liu]{zha2021douzero}
Zha, D., Xie, J., Ma, W., Zhang, S., Lian, X., Hu, X., and Liu, J.
\newblock Douzero: Mastering doudizhu with self-play deep reinforcement learning.
\newblock In \emph{International Conference on Machine Learning (ICML)}, pp.\  12333--12344, 2021.

\bibitem[Zhao et~al.(2022)Zhao, Zhao, Hu, Zhou, and Li]{zhao2022douzero+}
Zhao, Y., Zhao, J., Hu, X., Zhou, W., and Li, H.
\newblock Douzero+: Improving doudizhu ai by opponent modeling and coach-guided learning.
\newblock In \emph{IEEE Conference on Games (CoG)}, pp.\  127--134, 2022.

\end{thebibliography}
\bibliographystyle{icml2024}


\newpage
\appendix
\onecolumn
\section{Naruto Mobile}
\label{a_Naruto}
\subsection{Introduction of Naruto Mobile}
Naruto Mobile is an online fighting game published by Tencent Games with over 100 million registered users. Naruto Mobile has a large-scale character pool with more than 400 characters(ninjas). Each ninja has its special unique characteristics. Players of Naruto Mobile can choose a ninja from the character pool and use the ninja to fight against other characters. The winning condition for all players is to defeat their opponents. In every episode, two players choose a character of their own, and they control their character against the other player(i.e., the opponent). The episode will terminate when a player's health drops to 0 or the game time ends. Our objective is to create an AI system for Naruto Mobile that is seamlessly integrated into the game and actively participates in battles against players. This AI system will be consistently deployed, ensuring its continuous presence and engagement within the game environment. Besides, the system is required to have sufficient generalization capabilities to be able to control all characters through only one model.

Each ninja is controlled using a virtual joystick to manage movement and direction. They possess a basic attack, two primary skills, one ultimate skill, one summoning skill, and one defensive skill. Furthermore, certain skills have a mechanism that allows for multiple subsequent activations after their initial release. Additionally, ninjas have the option to choose different scroll items to enhance the diversity of their combat strategies. 
\subsection{Hitbox and Hurt Box}
The hurt box and hitbox play crucial roles in determining the outcome of attacks. The hurt box represents the vulnerable area around a character's body where they can take damage. It acts as an invisible or transparent box that detects incoming attacks. When an opponent's attack connects with the hurt box, the character will suffer damage. The size and shape of the hurt box can vary based on the character's stance, animation, and the specific move being executed.

On the other hand, the hitbox represents the area around a character's attacks where they can deal damage to the opponent. Similar to the hurt box, it is an invisible or transparent box that detects potential hits. When a hitbox overlaps with the hurt box of the opponent's character, it registers as a successful hit, causing damage. The size and shape of the hitbox can vary depending on the specific move being executed.

Both the hurt box and hitbox are integral components of the hit detection system in fighting games. They determine whether an attack successfully lands on an opponent, resulting in damage, or if it misses. Precise placement and sizing of the hitbox and hurt box are essential for balanced and fair gameplay, as they directly impact the effectiveness and reliability of attacks and defensive maneuvers.

\subsection{Different Information in Naruto Mobile}
\begin{figure}[htb]
    \centering
    \includegraphics{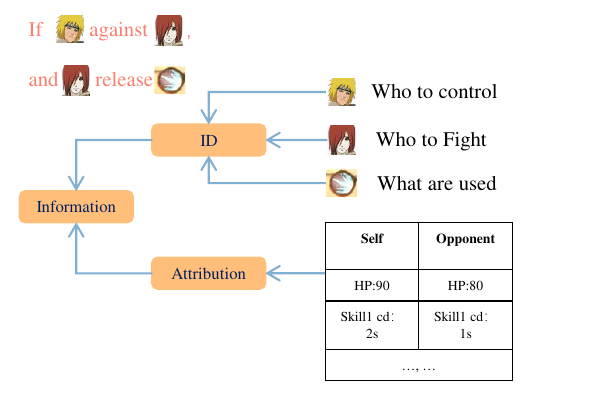}
    \caption{ID information and attribution information}
    \label{ID information and attribution information}
\end{figure}

All the information from the client is divided into distinguishable categories to model both self and opponents effectively(\cref{ID information and attribution information}). This differentiation involves the utilization of ID information, which provides directed references to characters, skills, actions, etc., and general numerical information, which represents less explicit inputs to the network.

From \cref{ID information and attribution information}, it was observed that the ID formation represents the figuration that which character is being controlled, which character is fighting with, and what skills are used. The attribution information is the specific numerical value of the entity.
\subsection{MDP of Naruto Mobile}
Considering the opponent as a component of the environment, the Markov Decision Process (MDP) framework for Naruto Mobile's training agent is formalized as a six-element tuple $\langle \mathcal{S}, \mathcal{A}, P, R, T, \gamma \rangle$. Here, $\mathcal{S}$ represents the state space and $\mathcal{A}$ denotes the action space. $R$ is defined as the reward function, $P$ encapsulates the state transition probabilities, $T$ signifies the time horizon, and $\gamma$ is the discount factor. At each discrete time step $t$, the agent observes the current state of the environment, denoted as $s_t \in \mathcal{S}$. The agent's policy $\pi$, a mapping from states to actions, selects an action $a_t$ based on the current state, formalized as $a_t \sim \pi(\cdot|s_t)$. The state then transitions to the next state $s_{t+1} \sim P(\cdot|s_t,a_t)$, and the agent receives a reward $r_{t+1}$. The state value function is $V_{\pi}(s)=\mathbb{E_{\pi}}[\sum\limits_{k=0}^{T}\gamma^{k}R_{t+k+1}|S_t=s]$, and the action value function is $Q_{\pi}(s)=\mathbb{E_{\pi}}[\sum\limits_{k=0}^{T}\gamma^{k}R_{t+k+1}|S_t=s, A_t=a]$ are always leveraged to optimize policy. The advantage function $A(s,a)=Q(s,a)-V(s)$ is used to measure the quality of action $a$ compared to the average quality.

The reward function $r_t$ is defined based on the variations in health points $r^{t}_{HP}$ and the game's outcome. $r_{t}^{HP}$ is defined as:
\begin{equation}
    r_{t}^{HP}={HP}_{t+1}^{self}-{HP}_{t}^{self}+{HP}_{t}^{oppo}-{HP}_{t}^{oppo},
\end{equation}
Where the ${HP}^{self}$ is the health points of the agent, and ${HP}^{oppo}$ is the health points of the opponent. Then the reward function $r_t$ is defined as:
\begin{equation}
    r_t=\left\{
    \begin{split}
        & r_{t}^{HP}, & t<T\\
        7&\cdot r_{t}^{HP}, & t=T\\
    \end{split}
    \right.
\end{equation}

The agent is constructed by a neural network with parameter $\theta$, i.e.,  the policy $\pi_{\theta}(a_t|s_t)$. The optimization goal of the RL algorithm is to find an optimal policy $\pi^*$ to maximize the cumulative reward:
\begin{equation}
    \pi^{*} = \arg\max\limits_{\pi_{\theta}}\mathbb{E}_{(s_t,a_t)\sim\pi_{\theta}}[\sum\limits_{t=0}^{n}\gamma^{t}r(s_{t}, a_{t})]
\end{equation}

\subsection{Action Representation of Naruto Mobile}
 The action space of Naruto Mobile is composed of three action heads, each represented by a one-hot vector. When predicting an action, the network simultaneously predicts all the action heads. These action heads are then passed through an action mask processing and delivered to the client for execution. \cref{table_action} shows the action head.

 \begin{table}
\centering
 \caption{Action representation of Naruto Mobile}
 \label{table_action}
\begin{tabular}{c|c|c|c|c}
\hline
Action Space & Action Head & Action Name &  & Key mapping \\ \hline
Move & Action\_ud & None &  & JoyStick (left hand) \\ \cline{3-5} 
 &  & Up &  &  \\ \cline{3-5} 
 &  & Down &  &  \\ \cline{2-5}
 & Action\_lr & None &  &  \\ \cline{3-5} 
 &  & Left &  &  \\ \cline{3-5} 
 &  & Right &  &  \\ \hline
Skill & Skill & None & Substitute & Fight buttons (right hand) \\ \cline{3-5} 
 &  & Punch & Summon &  \\ \cline{3-5} 
 &  & Skill1 & Scroll &  \\ \cline{3-5} 
 &  & Skill2 & Subskill1 &  \\ \cline{3-5} 
 &  & Skill3 & Subskill2 &  \\ \hline
Direction & Direction & Direction1 & Direction5 & Joystick skill \\ \cline{3-5} 
 &  & Direction2 & Direction6 &  \\ \cline{3-5} 
 &  & Direction3 & Direction7 &  \\ \cline{3-5} 
 &  & Direction4 & Direction8 &  \\ \hline
\end{tabular}
\end{table}

\section{Related Work}
\label{a_Related}
\subsection{Policy Improvement in Multi-agent Competition}
Multi-agent environment means there exists more than one agent in the environment. These agent needs to cooperate or compete to achieve their goals. In the competition setting, agents defeat other agents to win. Every agent has to improve their policy to make sure that is un-exploitable. There exists plenty of research to achieve un-exploitable. Naive self-play is a common approach to policy improvement, it essentially does best response in an iterated way. Self-play only focuses on finding the best response to the opponent's latest strategy, which would lead to circular behaviors in in-transitive games. To address the limit of self-play, Fictitious Play (FP)~\citep{brown1951iterative} was proposed. FP maintains a belief over the historical action that the opponent has played, the learning agent then takes the best response to this empirical distribution. Generalized Weakened Fictitious Play (GWFP)~\citep{leslie2006generalised} was proposed to weaken the constraint of FP, which was accomplished by allowing approximate best response and perturbed average strategy updates while preserving the convergence guarantee under certain conditions. 

The methods mentioned above are in the formulation of Norm-formal Game (NFG). To expand NFG to Extensive-form Game (EFG) while utilizing modern machine learning, two variants of FP are implemented by Fictitious Self-play (FSP) \citep{Heinrich2015FictitiousSI}. The first variant introduces full-width extensive-form fictitious (XFP). It is realization equivalent to a normal-form fictitious play and therefore inherits its convergence guarantees. The second variant is FSP based on the first variant but leveraging reinforcement learning (RL) to do the best response and supervised learning (SL) to average strategy update. With the development of deep learning,~\citet{heinrich2016deep} leverages the neural networks to implement the RL and SL of FSP. Prioritized Fictitious self-play (PFSP)~\citep{vinyals2019grandmaster} calculates weight based on winning rate and selects opponents based on weight to avoid the waste of computing resources.

Double Oracle (DO)~\citep{mcmahan2003planning} is also an iterated best response method, different from self-play, it best responds to the opponent's Nash equilibrium at each iteration. Although DO can guarantee the convergence, it will traverse the entire policy space in the worst case. Policy Space Response Oracle (PSRO)~\citep{lanctot2017unified} is a generalization of DO on meta-games. DO focuses on the action at each iteration, but PSRO focuses on policy space for every agent. Thus, PSRO can generalize all previous methods by setting different forms of the distribution of policy space.
\subsection{DRL Application in Games}
With the development of deep reinforcement learning (DRL), there are several successful applications in different domains such as robotic \citep{andrychowicz2017hindsight}, autonomous vehicles \citep{aradi2020survey}, and video games. Video games are the most suitable scene serving as DRL environment owing to their data accessibility and timely feedback. Leveraging DRL to develop game AI in video games has been stimulating researchers' interest.
DQN \citep{mnih2015human} proposed trying its hand at atari games using DRL from pixel input. Subsequently, AlphaGo \citep{silver2016mastering}, efficiently combined the policy and value networks with MCTS and defeated the world champion of Go. Without human knowledge, AlphaGo Zero \citep{silver2017mastering} defeated AlphaGo solely through self-play to improve policy. \citet{silver2018general} generalizes AlphaGo Zero to AlphaZero which can achieve superhuman competence in chess, shogi, and Go through self-play. The result of AlphaZero shows the power of DRL and self-play.  

After tackling the simple perfect-information game, researchers' interest began to focus on more complex game scenarios such as poker, Dota2, and StarCraft II. These environments have more complex state-action space. DeepStack \citep{moravvcik2017deepstack} defeated with statistical significance professional poker players in heads-up no-limit Texas hold 'em which combines recursive reasoning, decomposition, and learned intuition through self-play. DouZero \citep{zha2021douzero} and DouZero+ \citep{zhao2022douzero+} are proposed to solve DouDizhu through enhancing traditional Monte-Carlo methods with deep neural networks, action encoding, parallel actors, and opponent modeling. For real-time strategy (RTS) games, AlphaStar \citep{vinyals2019grandmaster} achieved the grandmaster level in StarCraft II through league training with PFSP. And for the sub-genre of RTS games, Multi-player Online Battle Arena (MOBA) games are also challenging. OpenAI Five \citep{berner2019dota} is the first AI system to defeat the world champions in a famous MOBA game as known as Dota2. OpenAI Five leveraged existing reinforcement learning techniques, scaled to learn from batches of approximately 2 million frames every 2 seconds. \citet{ye2020mastering} tackled MOBA game from the perspectives of both system and algorithm. They further proposed a MOBA AI learning paradigm that methodologically enables playing full MOBA games with deep reinforcement learning \citep{ye2020towards}.

In this paper, we focus on the Fighting game which is an imperfect-information two-player zero-sum game called Naruto Mobile. Naruto Mobile has a large pool of characters and complex state-action space, with a high demand for real-time feedback. We design an AI system with heterogeneous league training to tackle these challenges. Our system trained on only 15\% of the characters but achieves high-level player competence across all characters. In particular, our system is consistently deployed in the game and engages in battles with players.

\subsection{Generalization in Reinforcement Learning}
Previous research has extensively investigated the challenges of generalization in reinforcement learning. One line of work focuses on addressing the problem of overfitting, where an RL agent fails to generalize its learned policy to unseen states or tasks. Approaches such as experience replay~\citep{mnih2015human} and regularization techniques~\citep{houthooft2016vime} have been proposed to mitigate this issue. Another direction of research explores the concept of transfer learning in RL, aiming to enable knowledge transfer from one task to another. Methods like domain adaptation~\citep{sun2017learning} and meta-learning~\citep{finn2017model,lan2024contrastive} have been employed to facilitate better generalization across tasks. 
\citet{DBLP:conf/cvpr/HeSZH23,DBLP:journals/corr/abs-2404-12754}~discusses the capacity loss problem in reinforcement learning to ensure better stability and generalization of the network. Additionally, recent studies have investigated the role of model-based RL algorithms~\citep{hafner2019learning} and intrinsic motivation~\citep{pathak2017curiosity} in enhancing generalization capabilities. \citet{li2023normalization} and \citet{lyu2024cross,lyu2024towards,lyu2024understanding} discuss the generalization caused by vector input and vision input.There are some methods that employ causal inference and language models to enhance the multi-task generalization of DRL~\citep{Peng2023ConceptualRL,Peng2023SelfdrivenGL}, and some object-oriented generalization works~\citep{DBLP:conf/nips/Yi0PG0Dz0C22,DBLP:conf/icml/Yi0PGGYCLHDZGC23}. In addition, there is also some work on network pruning that can be considered for enhancing the generalization of models in reinforcement learning~\cite{wang2023brave}. Despite these efforts, the problem of achieving robust and reliable generalization in RL remains a challenging and active research area.

\section{Details about HELT}
\subsection{Learning Architecture}
\label{a_learning_arch}
\begin{figure*}[htb]
    \centering
    \includegraphics[width=\textwidth]{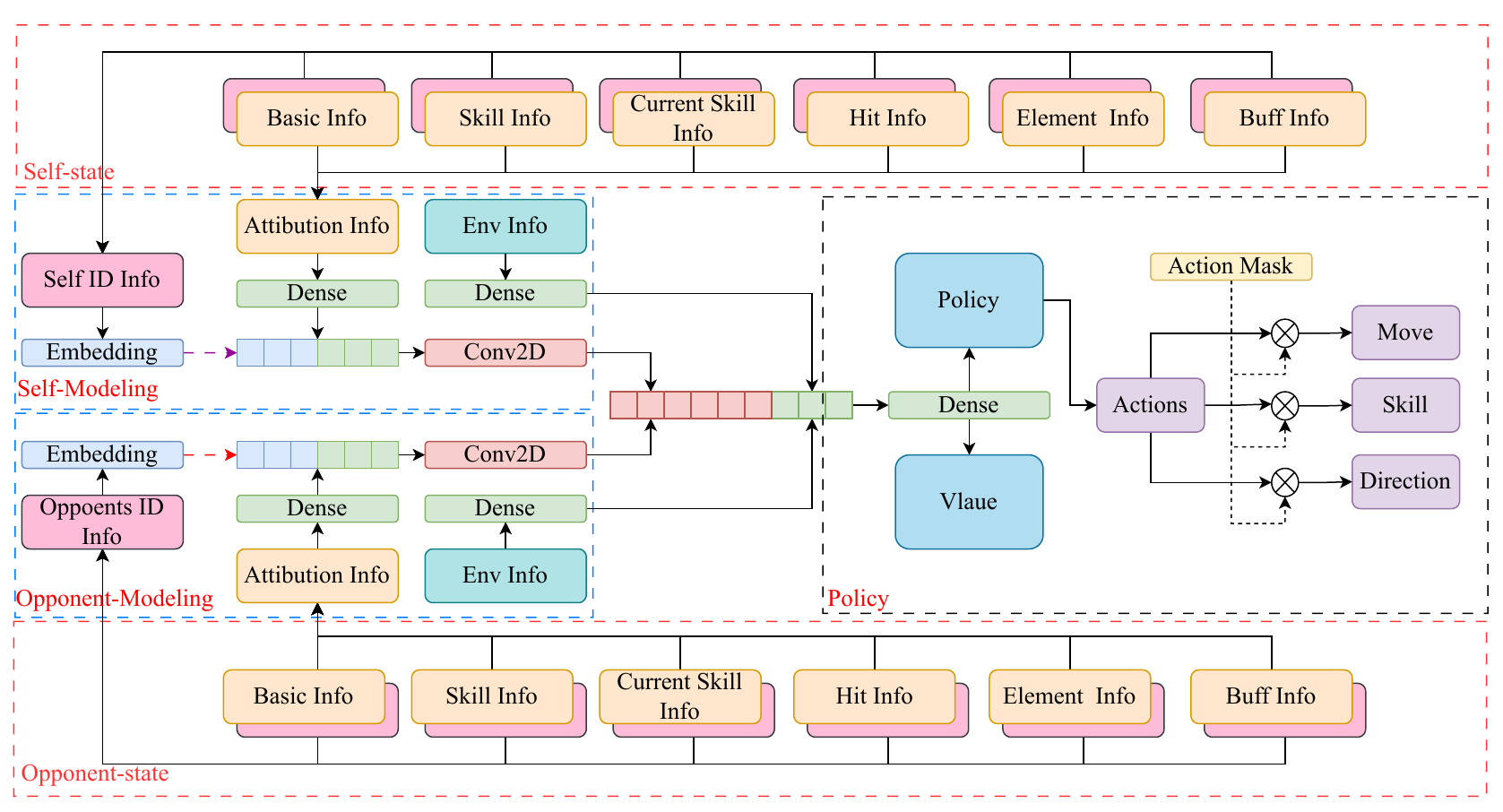}
    \caption{The structure of FIS, QS, and FQS. while the purple dashed line and the red dashed line are activated, the structure is FIS. When the purple dashed line is activated and the red dashed line deactivated, the structure is QS. While the purple dashed line and the red dashed line are both deactivated, the structure is FQS.}
    \label{a_fig_arche}
\end{figure*}

\cref{a_fig_arche} shows the learning architecture of Naruto Mobile. The input of the QS network comprises two components: self-modeling, and opponent-modeling. Self-modeling still utilizes the ID information and attribution information. Opponent modeling of the QS network only with attribution information, thereby liberating the QS network from the constraints of the opponent's character pool and enhancing the generalizability of the QS network. 

Env information incorporates spatial and relative characteristics of the two agents, in addition to their health and time attributes. All of these features are encoded through fully connected layers and concatenated. The agent's actions, encompassing movement direction and optional skills, are generated through the policy network. The purpose of the QS network is to improve generalization while ensuring strength. The QS agent is expected to be able to serve as a unified model to control a part of characters against all characters.

The FQS network is similar to the QS network and does not use any ID information, only the attributes information. FQS agent is expected to obtain models that can be used across all characters.

Different from the QS and FQS networks, FIS uses full ID information to model all the entities whether self-modeling or opponent-modeling. The advantage is that FIS can find the optimal model to defeat the opponent. The disadvantage is that when facing an opponent FIS has never seen before, the competence will decrease.

\subsection{Opponent Policy Selecting}
\label{a_oppo_select}
\textbf{Prioritized fictitious self-play(PFSP)} 
In this paper, our PFSP algorithm is similar to AlphaStar. Giving a learning agent $A$, we sample the frozen opponent $O$ from a candidate set $\mathbf{S}$ with probability
\begin{equation}
    \frac{f(\mathbb{P}[A\ beats\ O])}{\sum_{S\in\mathbf{S}}f(\mathbb{P}[A\ beats\ S])}
\end{equation}
Where $f:[0,1]\rightarrow[0,\infty]$ is the weighting function.

In this paper, we apply $f_{hard}(x)=(1-x)^p$ as the hard weighting to encourage focusing on the hardest opponents; and $f_{var}(x)=x(1-x)$ as the var weighting to encourage focusing on the opponents around the same level. Note that not only do we dynamically utilize PFSP based on win rates during the training process, but we also employ PFSP weights to streamline the model pool when determining opponents for each iteration.

\subsection{Populating the League}
Same as Alphastar, throughout the training process, we employed three distinct types of agents that primarily differed in their opponent distribution during training, the snapshotting procedure used to create a new player, and the probability of resetting to the previous parameters. Due to the lack of high-quality player data in Nurotu Mobile, we have designed different resetting mechanisms to optimize our exploitation.

\begin{figure}
    \centering
    \includegraphics[]{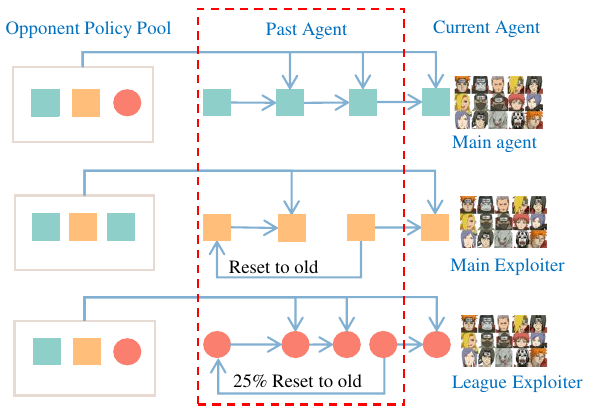}
    \caption{Population the league.}
    \label{fig_league_Training}
\end{figure}

Main agents are trained with a proportion of 100\% PFSP against all past players in the league. Although Alphastar leverages different proportions of PSFP and SP, applying heterogeneous agents ensures diversity among agents within the entire league. When timeout hours or the main agent defeats all agents in the league in more than 80\% of games, a copy of the main agent is added as a new agent to the league. The main agent never reset.

League exploiters are trained with PFSP and their frozen copies are added to the league when they beat all agents in the league in more than 80\% of games or after a timeout. At the beginning of the next iteration, the league exploiters have a 25\% probability that the agent is reset. Owing to the lack of high-quality player data, the league exploiter randomly selects a model from the models of the previous three iterations to serve as the current model. This mechanism allows the league exploiter to balance convergence speed and the ability to explore diversity effectively. The league exploiter has a reset probability of 0 for the first three generations of training.

Main exploiters always play against main agents. To address the issue of low win rates during the initial stages of training for the main exploiters, they are trained by PFSP with $f_{var}$ in the first three iterations. Main agents are added to the league when the main agents are defeated in more than 80\% of games, or after a timeout. The reset mechanism is similar to the league exploiter, but the main exploiter will reset at every start of the iteration. The primary objective of main exploiters is to identify and expose weaknesses within the main agents, thereby facilitating their enhancement and fortification.

\subsection{Opponent Character Selecting}
The opponent character selecting strategy is accomplished by \citet{wang2024balancing}. The idea is based on regret matching (RM). For details, readers can refer \cite{wang2024balancing}.

Given a match between $i$ and $j$, the return to $j$ is defined as:

\[
    r_{s_t}(i, j)= \begin{cases}
        1, j\ \text{wins} \\
        0, \text{otherwise} \\
    \end{cases}
\]

Once the result of the match comes out, exponential smoothing is applied to the return value and this averaged result serves as an updated win rate:

\[
    p_{s_t}(win_{i>j}) = \bar{r}_{s_t}(i, j) = \bar{r}_{s_t-1}(i, j) * \gamma + r_{s_t}(i, j) * (1-\gamma)
\]

where $\gamma\in [0, 1]$ is the smoothing factor.

Then the expected utility is defined as the overall win rate, which can be calculated as: 
\[
\mathbb{E}(p_{s_t}) = p_{s_t}(win_{all}) = \sum_{i,j\in I}p_{s_t}(win_{i>j}) w_{s_t-1}(i,j)
\]

The regret is defined as the difference between the smoothed win rate and the expected win rate:
\[
    \Delta p_{s_t}(i, j) = \bar{r}_{s_t}(i, j) - \mathbb{E}(p_{s_t})
\]
and the regret matrix is updated as:
\[
    R_{s_t}(i, j) = \max(R_{s_t-1}(i, j) + \Delta p_{s_t}(i, j), 0)
\]
Finally, new weights for selecting combinations in the next match are updated as:
\[
w_{s_t}(i,j) = \begin{cases}
    \frac{1}{N^2}, \text{if}\ \sum_{i, j \in I} R_{s_t}(i,j) =0, \\
    \frac{R_{s_t}(i, j)}{\sum_{i, j \in I} R_{s_t}(i,j)} * (1-\eta) + \frac{1}{N^2} * \eta \  \text{otherwise} \\
\end{cases}
\]
where $\eta$ is a weight factor.

\subsection{Policy Improvement}
\label{a_policy}
In \cref{markv}, by modeling the fighting game as a Markov process, we can leverage reinforcement learning algorithms to optimize the agent's strategies. We use the Proximal Policy Optimization (PPO) algorithm \citep{schulman2017proximal} in our system. PPO trains a value function $V_{\phi}(s_t)$ with a policy $\pi_{\theta}(a_t|s_t)$, the utilization of importance sampling makes it highly feasible to apply reinforcement learning algorithms in distributed scenarios.

\textbf{Policy updates.} In our large-scale distributed system, the trajectories are sampled from multiple sources of policies, which can exhibit significant deviations from the current policy $\pi_{\theta}$. While the method proposed by \citep{ye2020mastering} demonstrates superior competence in MOBA games, we have discovered that the original Proximal Policy Optimization (PPO) algorithm is more effective for fighting games. Hence, our update objectives have remained consistent with PPO:
\begin{equation}
\begin{split}
    \mathcal{L}^{\pi}(\theta)=\mathbb{E}_{t}[min(\rho_{t}(\theta)\hat{A_t}, clip(\rho_t(\theta),1-\epsilon,1+\epsilon)\hat{A_t})]+\beta\mathcal{L}^{E}(\theta)
\end{split}
\end{equation}
Where $\rho_t(\theta)=\frac{\pi_{\theta}(s_t|a_t)}{\pi_{old}(s_t|a_t)}$ is the important weighting and $\hat{A_t}$ is the advantage estimation that is calculated by Generalized Advantage Estimation (GAE) \citep{schulman2015high}. $\mathcal{L}^{E}$ is the entropy of policy which enhances the exploration. 

\textbf{Value updates.} Same as AlphaStar, we use full information to decrease the variance of value estimation:
\begin{equation}
    V^{\pi_{\theta}}(s_t)=\mathbb{E}_{a\sim\pi_{\theta}}[r(s_t, a)+\gamma V^{\pi_{\theta}}(s_{t+1})]
\end{equation}
We use GAE to estimate the advantage function $\hat{A}$:
\begin{equation}
\begin{split}
     A^{\pi_{\theta}}(s_t, a_t)=&r(s_t,a_t)+\gamma V^{\pi_{\theta}}(s_t+1)-V^{\pi_{\theta}}(S_t),\\
    & \hat{A}=\sum\limits_{t=0}^{T}{\gamma\lambda}^{t}A_{t+1},
\end{split}
\end{equation}
and the value loss can be expressed as follows:
\begin{equation}
    \mathcal{L}^{V}(\phi)=\mathbb{E}_{a\sim\pi_{\theta}}[\sum\limits_{k=t}^{T}\gamma^{k-t}r(s_k,a_k)-V(s_t)]^2.
\end{equation}

\section{Experimental Details}
\label{a_experi}
\subsection{Experimental Setting}
In \cref{table1}, we list the parameters used in the experiments. All the parameters are the same for three heterogeneous agents. During the training process, the main agent, league exploiter, and main exploiter are initialized with different structures. These three types of league agents maintain the same policy pool. After each agent completes an iteration, its copy is added to the overall policy pool. We evaluate the policies in the policy pool pairwise, and based on the evaluation results, we use the PFSP to select the next round of strategies from the policy pool. Main exploiters will reset at every start of the iteration and league exploiters will reset at the same moment with a probability of 25\%. The main agents never reset. The three league agents are trained asynchronously, and each agent will undergo 10 training iterations. The maximum duration for each iteration is limited to 12 hours. 

\subsection{The Characters Selecting of Training Subset}
In Naruto Mobile, characters are categorized into ranks (S, A, B, C) based on their relative strength and skill mechanics by professional human experts. Taking this ranking system into consideration, higher-ranked characters are generally more powerful than lower-ranked ones. Building upon this foundation, we further selected characters suitable for our training based on their popularity within each rank, using expert priors. We will include these details in the appendix section of our paper for easy reference and further discussion.

\begin{table}
\centering
\caption{Eperimental Parameters}
\label{table1}
\begin{tblr}{
  vline{3} = {-}{},
  hline{1,7} = {-}{0.08em},
  hline{2} = {-}{},
}
Parameter                  & Value & Parameter            & Value \\
n-steps                    & 100   & Batch~ size          & 5120  \\
$\gamma$                   & 0.995 & Actor number         & 1000  \\
$\lambda$ & 0.95  & Env number per actor & 10    \\
Learning rate              & 2e-4  & Learner number       & 2     \\
CPU core num               & 9000  & GPU per Learner      & 0.5   
\end{tblr}
\end{table}
\subsection{Ablation Study}
We need to do ablation studies to figure out where the HELT's competence and generalization ability come from. For the generalization ability, we can take another look at \cref{performance} and \cref{generalization}, we can observe that while the ID-based FIS structure can lead to competence improvements, quantifying the states can enhance generalization, especially when it comes to opponent modeling. By quantifying opponent modeling while ensuring competence, we can improve the overall generalization capability.
\begin{figure*}[htb]
	\centering
	\subfigure[Impact of different league agents.] {\includegraphics[width=.3\textwidth]{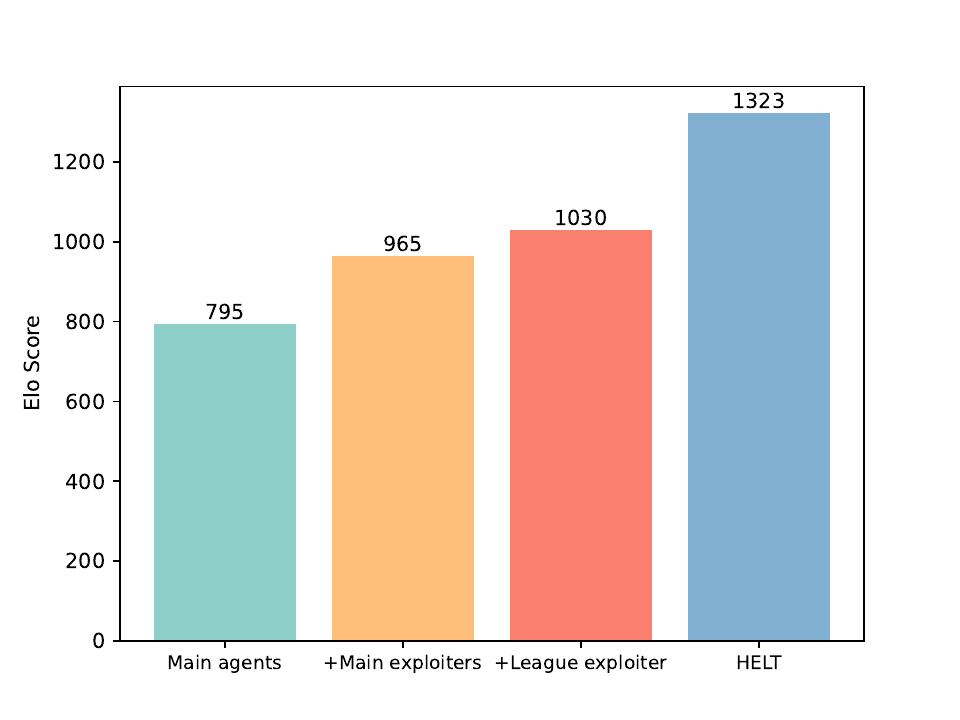}\label{ab_QS}}
	\subfigure[Impact of different opponent policy select strategies. ] {\includegraphics[width=.3\textwidth]{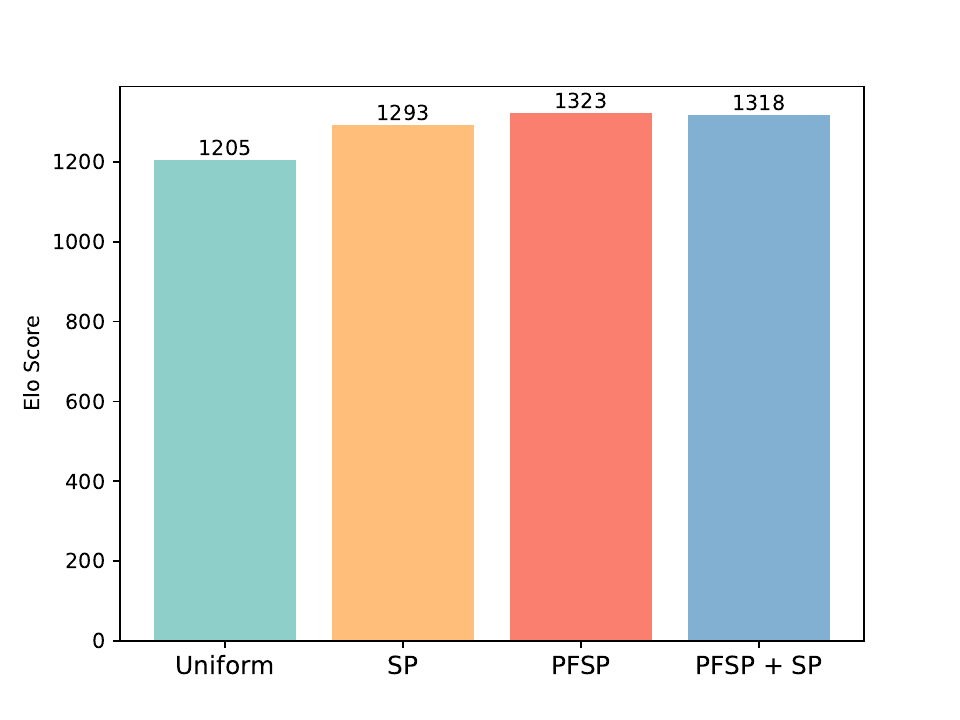}\label{policy_qs}}
	\subfigure[Impact of different character select  strategies] {\includegraphics[width=.3\textwidth]{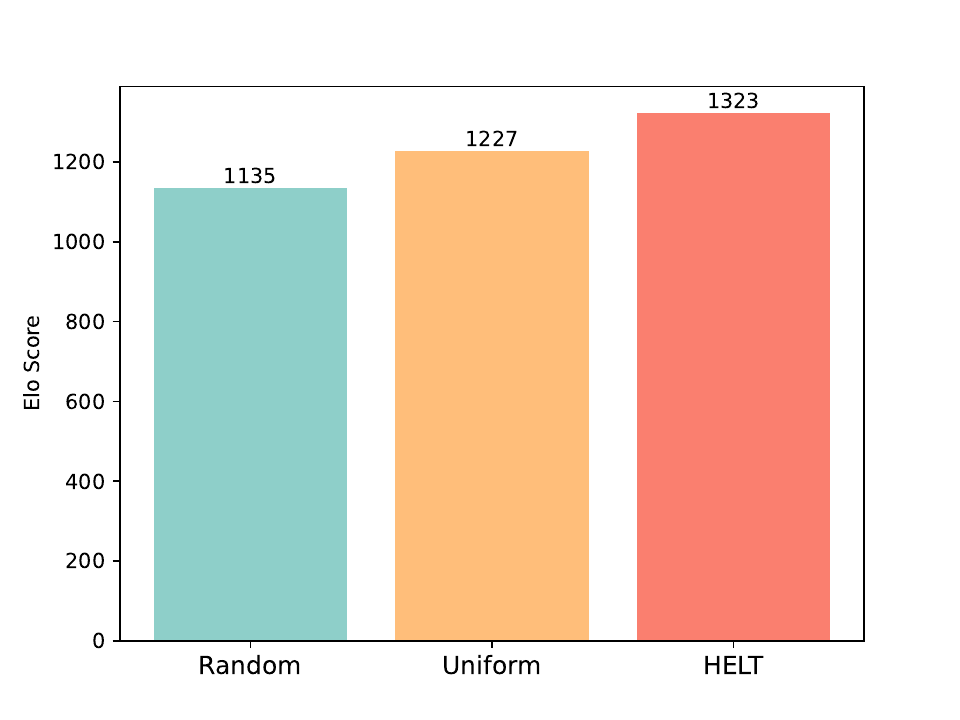}\label{char_qs}}
	\caption{Ablation studying of different components for QS agent. (a) Impact of different league agents. (b) Impact of different opponent policy select strategies. (c) Impact of different character select  strategies}
	\label{fig_ablation}
\end{figure*}

For competence, we conducted ablation experiments on different components using the QS agent as the focus of our study. This analysis aimed to observe the competence improvements brought by each component. \cref{ab_QS} displays the impact of different league agents on the model's competence. We can observe that although the same-structure league exploiter contributes to around a 20\% competence improvement, the primary source of competence in HELT comes from the league exploiter with the FIS structure. The inclusion of the FIS agent increases the overall diversity of strategies, significantly enhancing the competence of the QS agent. 
\cref{policy_qs} showcases the impact of different opponent policy selection strategies on the competence of the QS agent. From the \cref{policy_qs}, we can observe that the competence differences caused by different opponent policy selection methods are not significant. The strategy used in AlphaStar, PFSP+SP, shows similar competence to pure PFSP in our environment. Therefore, we primarily adopt the PFSP opponent policy selection in our study.

\cref{char_qs} displays the impact of different character selection strategies on the competence of the QS agent. We compared the strategies of random selection, uniform selection, and HELT's method\cref{a_oppo_select}. HELT's method encourages stronger characters to be trained more frequently, which results in the highest competence improvement.

Based on the above analysis, HELT's method can improve generalization while maintaining competence, which are both crucial factors for practical implementation and online gaming.

\subsection{Evaluating in Publication Platform}
\textit{Shūkai} primarily focuses on catering to the needs of commercial fighting games by aligning with player expectations, enhancing game content, and driving continuous game development. 
HELT is primarily developed to tackle the challenges of generalization and training efficiency that arise from the large character pool in commercial games. With its expansive roster of over 400 characters, Naruto Mobile is more challenging than FightingICE~\cite{khan2022darefightingice}, which features only 4 characters. 
As a result, our paper significantly focuses on deploying reinforcement learning systems in commercial fighting games. In contrast, FightingICE~\cite{khan2022darefightingice} places more emphasis on competition, aiming to ensure the generation of stronger agents.
Nevertheless, we still plan to evaluate HELT on the FightingICE platform to demonstrate its computational efficiency and generalization. However, since the framework we used is primarily tailored for game customization, we need to invest more time in integrating with FightingICE. We will release the performance of HELT on FightingICE in the future.

\section{Agent-Human Alignment with Extra Metrics}
\label{a_human_align}
\vskip 0.1in
\begin{figure*}[htb] 
\centering 
\subfigure[Error Rate]                                        {\includegraphics[width=.2\textwidth]{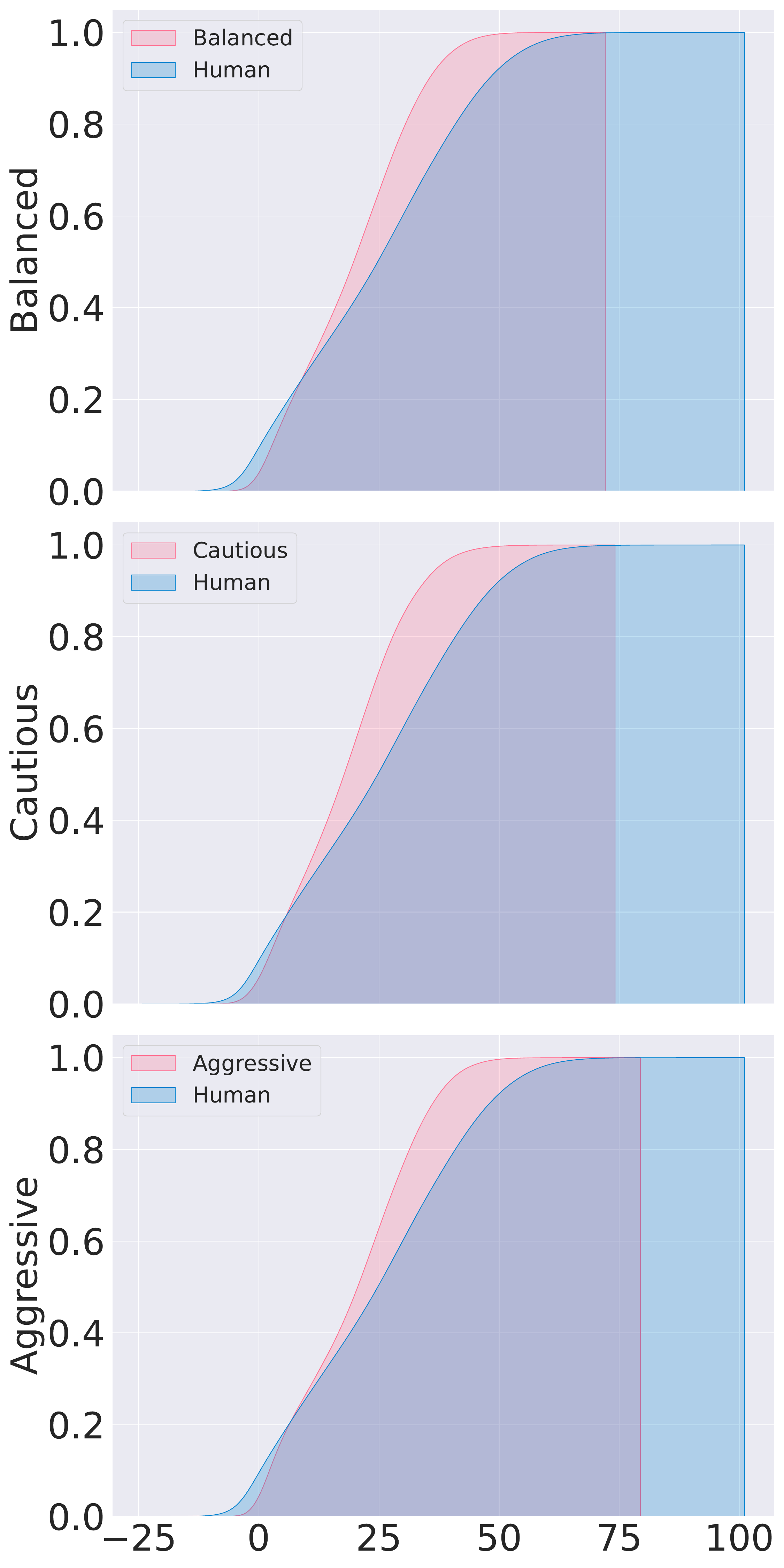}\label{Error Rate}} \subfigure[Skill1] {\includegraphics[width=.2\textwidth]{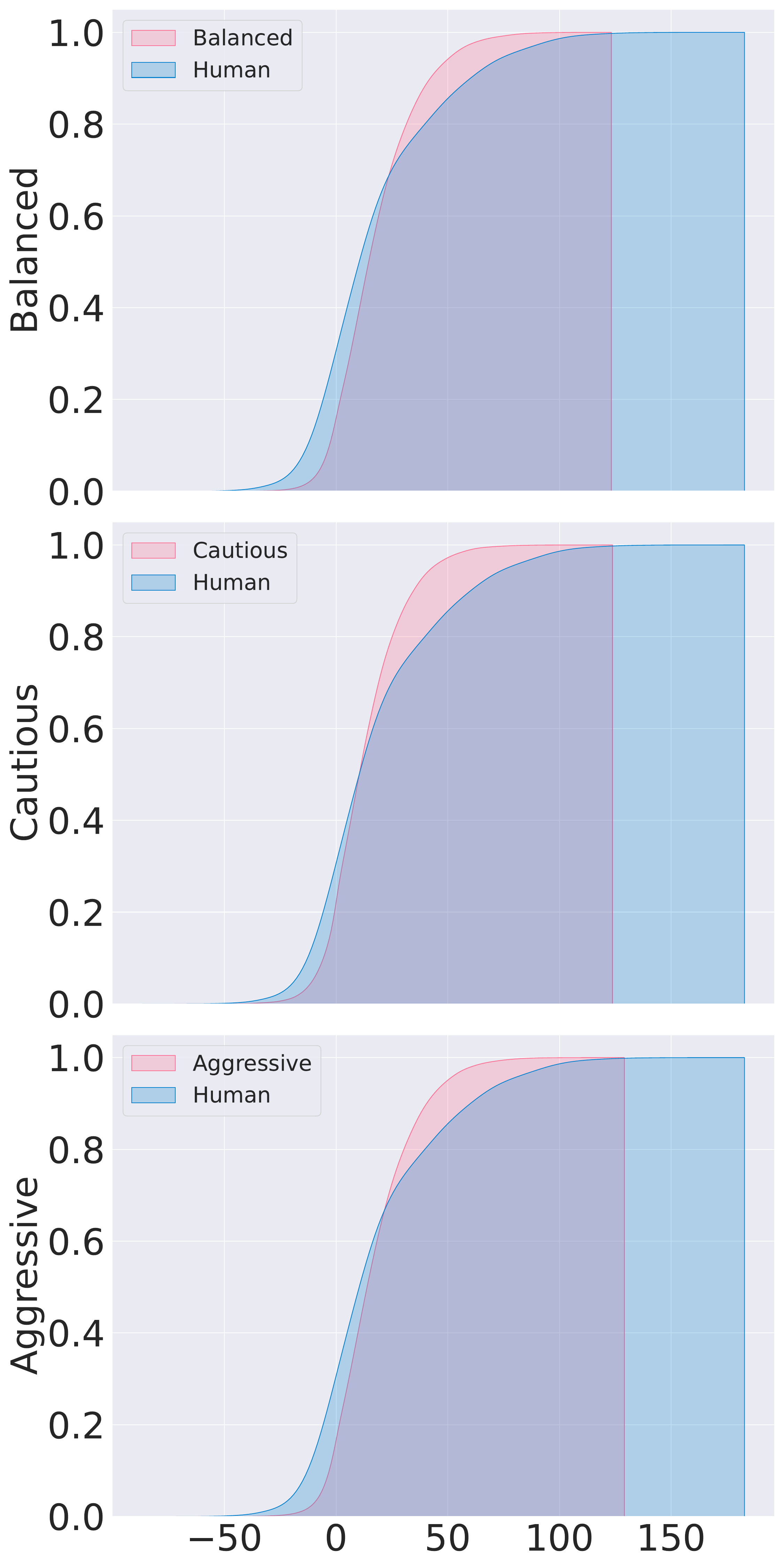}\label{Skill1}} \subfigure[Skill2] {\includegraphics[width=.2\textwidth]{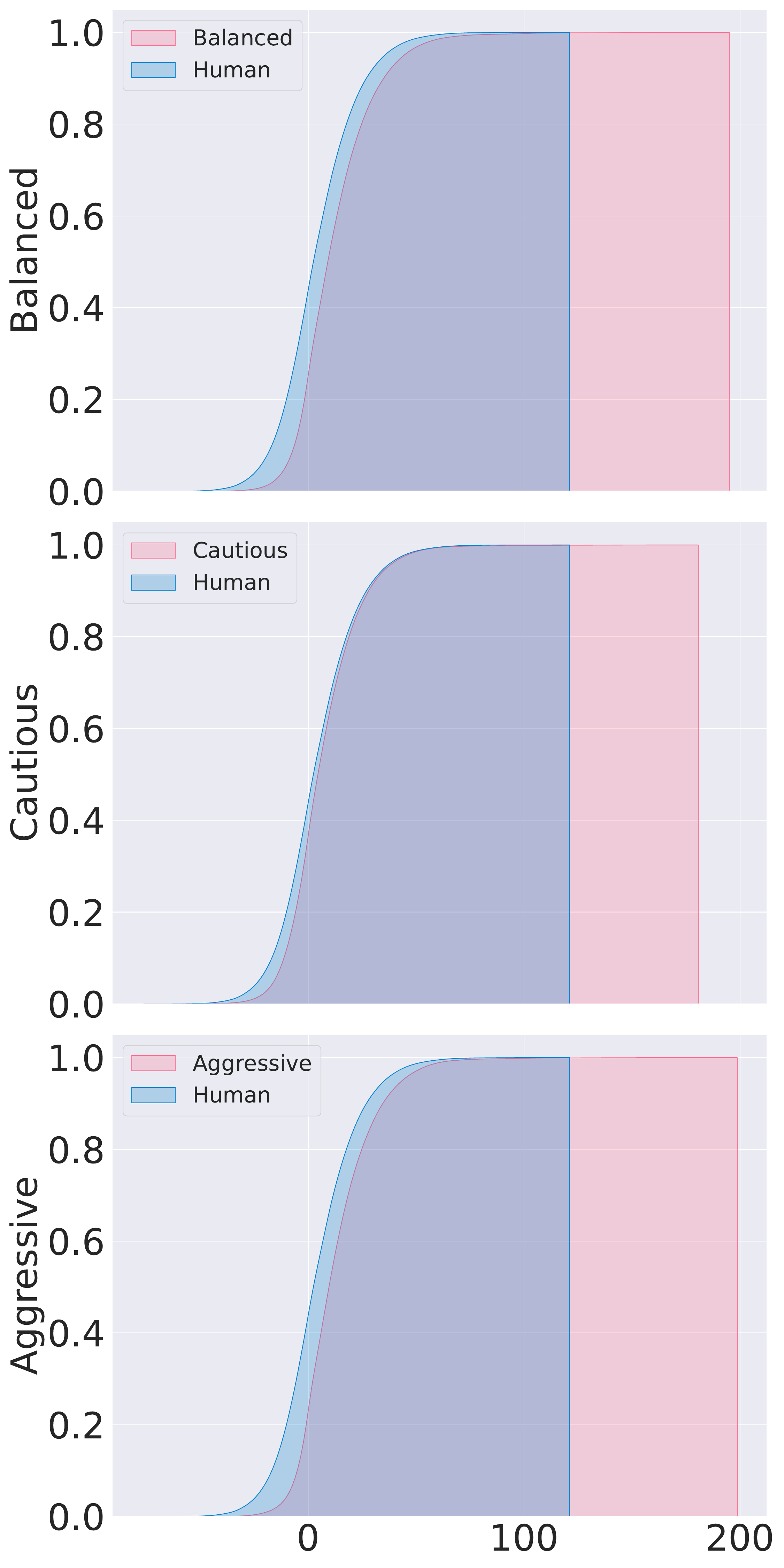}\label{Skill2}} \subfigure[Summon] {\includegraphics[width=.2\textwidth]{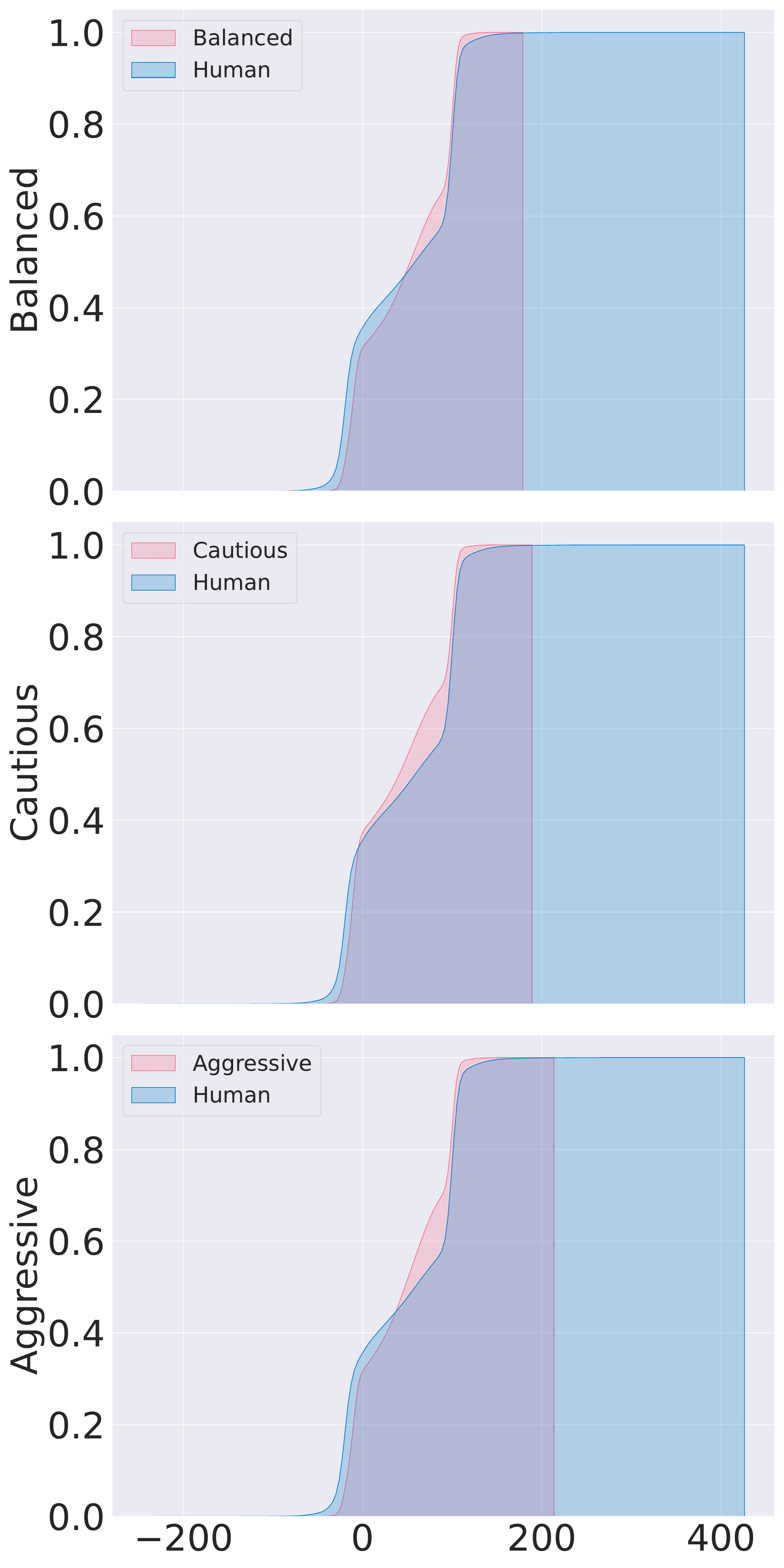}\label{PSY}} 
\caption{A comparison of the cumulative distribution function (CDF) between human players and different agents across different metrics.} 
\label{fig_distri_1} 
\end{figure*}
The error rate refers to the probability of making an error or failing to successfully perform a particular action or skill in a game. In the context of using substitution in Naruto Mobile, the error rate would indicate the likelihood of failing to execute the substitution properly.

Based on the previous discussion, it has been observed that our intelligent agents exhibit fewer errors compared to human players. To further analyze this, we compared the release of Skill 1 and Skill 2 between human players and the intelligent agents, considering the same character.


\end{document}